\documentclass{clv3}
\usepackage{url}
\usepackage{multirow}
\usepackage{subfigure}
\usepackage{times}
\usepackage{latexsym}
\usepackage{times}
\usepackage{stfloats}
\usepackage{url}
\usepackage{latexsym}
\usepackage{times}
\usepackage{float}
\usepackage{url}
\usepackage{graphicx}
\usepackage{epstopdf}
\usepackage{amsfonts}
\usepackage{amssymb}
\usepackage{fixltx2e}
\usepackage{amsmath}
\usepackage{verbatim}
\usepackage{blindtext}
\usepackage{multirow}
\usepackage{subfigure}
\usepackage{pifont}
\usepackage{amssymb}
\usepackage[algo2e]{algorithm2e}

\newcommand{\xmark}{\ding{55}}%
\issue{1}{1}{2006}

\begin{document}
	
	\newcommand{\thickhline}{\noalign{\hrule height 1pt}}
	\runningtitle{A Sequential Matching Framework for Multi-turn Response Selection in Retrieval-based Chatbots}
	
	\runningauthor{Wu et al.}

	\title{A Sequential Matching Framework for Multi-turn Response Selection in Retrieval-based Chatbots}
	
	\author{Yu Wu\thanks{Emails: \{wuyu,lizj\}@buaa.edu.cn  ~~~~  \{wuwei,v-chxing,can.xu,mingzhou\}@micorosoft.com}}
	\affil{Beihang University}
	
	\author{Wei Wu}
	\affil{Microsoft Research}
	
	\author{Chen Xing}
	\affil{NanKai University}
	
	\author{Can Xu}
	\affil{Microsoft Research}
	
	\author{Zhoujun Li}
	\affil{Beihang University}
	
	\author{Ming Zhou}
	\affil{Microsoft Research}
	
	\maketitle
	\begin{abstract}
		We study the problem of response selection for multi-turn conversation in retrieval-based chatbots.  The task requires matching a response candidate with a conversation context, whose challenges include how to recognize important parts of the context, and how to model the relationships among utterances in the context.  Existing matching methods may lose important information in contexts as we can interpret them with a unified framework in which contexts are transformed to fixed-length vectors without any interaction with responses before matching. The analysis motivates us to propose a new matching framework that can sufficiently carry the important information in contexts to matching and model the relationships among utterances at the same time. The new framework, which we call a sequential matching framework (SMF), lets each utterance in a context interacts with a response candidate at the first step and transforms the pair to a matching vector. The matching vectors are then accumulated following the order of the utterances in the context with a recurrent neural network (RNN) which models the relationships among the utterances.  The context-response matching is finally calculated with the hidden states of the RNN.  Under SMF, we propose a sequential convolutional network and sequential attention network and conduct experiments on two public data sets to test their performance. Experimental results show that both models can significantly outperform the state-of-the-art matching methods. We also show that the models are interpretable with visualizations that provide us insights on how they capture and leverage the important information in contexts for matching.

	\end{abstract}
	
	\section{Introduction}
	
	Recent years have witnessed a surge of interest on building conversational agents in both industry and academia.  Existing conversational agents can be categorized into task-oriented dialog systems and non-task-oriented chatbots. Dialog systems focus on helping people complete specific tasks in vertical domains \cite{young2010hidden}, such as flight booking, bus route enquiry, and restaurant recommendation, etc.; while chatbots aim to naturally and meaningfully converse with humans on open domain topics \cite{ritter2011data}. Building an open domain chatbot is challenging, because it requires the conversational engine to be capable of responding to any input from humans that covers a wide range of topics. To address the problem, researchers have considered leveraging the large amount of conversation data available on the internet, and proposed generation-based methods \cite{DBLP:conf/acl/ShangLL15,serban2015building,vinyals2015neural,li2016persona,xing2016topic,mou2016sequence} and retrieval-based methods \cite{wang2013dataset,hu2014convolutional,ji2014information,wang2015syntax,zhou2016multi,DBLP:conf/sigir/YanSW16}. Generation-based methods generate responses with natural language generation models learnt from the conversation data, while retrieval-based methods re-use the existing responses by selecting proper ones from an index of the conversation data. In this work, we study the problem of response selection in retrieval-based chatbots, because retrieval-based chatbots have the advantage of returning informative and fluent responses. While most existing work on retrieval-based chatbots studies response selection for single-turn conversation \cite{wang2013dataset} in which conversation history is ignored, we study the problem in a multi-turn scenario. In a chatbot, multi-turn response selection takes a message and utterances in its previous turns as an input and selects a response that is natural and relevant to the entire context.
	
	\begin{table}[!t]
		\caption{An example of multi-turn conversation}	
		\label{example1} \small
		\centering
		\begin{tabular}{l}
			\hline
			\textbf{Context} \\ \hline
			\emph{Human}: How are you doing? \\ \hline
			\emph{ChatBot}: I am going to \textbf{hold a drum class} in Shanghai. Anyone wants to join? The location \\is near Lujiazui.\\ \hline
			\emph{Human}: Interesting! Do you have coaches who can help me practice \textbf{drum}? \\ \hline
			\emph{ChatBot}: Of course. \\ \hline
			\emph{Human}: Can I have a free first lesson?\\ \hline
			\textbf{Response Candidates} \\ \hline
			\emph{Response 1}: Sure. Have you ever played drum before? \checkmark  \\ \hline
			\emph{Response 2}: What lessons do you want? \xmark \\ \hline
			
		\end{tabular}
		
	\end{table}
	A key step in response selection is measuring matching degree between an input and response candidates. Different from single-turn conversation in which the input is a single utterance (i.e., the message), multi-turn conversation requires context-response matching where both the current message and the utterances in its previous turns should be taken into consideration.  The challenges of the task include (1) how to extract important information (words, phrases, and sentences) from the context and leverage the information in matching; and (2) how to model relationships and dependencies among the utterances in the context. Table \ref{example1} uses an example to illustrate the challenges.  First, to find a proper response for the context, the chatbot must know that ``hold a drum class'' and ``drum'' are important points. Without them, it may return a response relevant to the message (i.e., the last turn in the context) but nonsense under the context (e.g., ``what lessons do you want?'').  On the other hand, words like``Shanghai'' and ``Lujiazui'' are less useful and even noisy to response selection. The responses from the chatbot may drift to the topic of ``Shanghai'' if the chatbot pays much attention to these words.  Therefore, it is crucial yet non-trivial to let the chatbot understand the important points in the context and leverage them in matching and at the same time circumvent the noise. Second, there is a clear dependency between the message and the second turn in the context, and the order of the utterances matters in response selection because there will be different proper responses if we exchange the third turn and the last turn.
	
	Existing work, including the recurrent neural network architectures proposed by Lowe et al. \cite{lowe2015ubuntu}, the deep learning to respond architecture proposed by Yan et al. \cite{DBLP:conf/sigir/YanSW16}, and the multi-view architecture proposed by Zhou et al. \cite{zhou2016multi} may lose important information in context-response matching because they follow the same paradigm to perform matching which suffers clear drawbacks. In fact, although these models have different structures, they can be interpreted with a unified framework: a context and a response are first individually represented as vectors, and then their matching score is computed with the vectors. The context representation includes two layers. The first layer represent utterances in the context, and the second layer takes the output of the first layer as an input and represents the entire context. The existing work differs in how they design the context representation and the response representation and how they calculate the matching score with the two representations. The framework view unifies the existing models and indicates the common drawbacks they have: everything in the context are compressed to one or more fixed-length vectors before matching is conducted; and there is no interaction between the context and the response in the formation of their representations.  The context is represented without enough supervision from the response, and so is the response.

	To overcome the drawbacks suffered by the existing work, we propose a sequential matching framework (SMF) that can tackle the two challenges of context-response matching simultaneously. SMF matches each utterance in the context with the response at the first step and forms a sequence of matching vectors. It then accumulates the matching vectors of utterance-response pairs in the chronological order of the utterances. The final context-response matching score is calculated with the accumulation of pair matching. Different from the existing framework, SMF allows utterances in the context and the response to interact with each other at the very beginning, and thus important matching information in each utterance-response pair can be sufficiently preserved and carried to the final matching score. Moreover, relationships and dependencies among utterances are modeled in a matching fashion, so the order of the utterances can supervise the aggregation of the utterance-response matching. Specifically, SMF consists of three layers. The first layer extracts important matching information from each utterance-response pair and transforms the information into a matching vector. The matching vectors are then uploaded to the second layer where a recurrent neural network with gated recurrent units (GRU) \cite{chung2014empirical} is employed to model the relationships and dependencies among the utterances and accumulate the matching vectors into its hidden states. The final layer takes the hidden states of the GRU as an input and calculates a matching score for the context and the response.
	
	The key to the success of SMF lies in how to design the utterance-response matching layer which requires identification of important parts in each utterance. We propose implementing the layer with a convolution-pooling technique and an attention technique, which results in a sequential convolutional network (SCN) and a sequential attention network (SAN) under SMF.  Specifically, given an utterance-response pair, SCN first constructs a word-word similarity matrix and a sequence-sequence similarity matrix by embedding of words and hidden states of a GRU on the sequence of words respectively.  The two matrices, which represent utterance-response matching on a word level and a segment level \footnote{Here a segment represents a subsequence of an utterance} respectively, are then transformed and fused as a matching vector through an alternation of convolution and pooling operations. In SCN, the response helps recognize the important words or segments in each utterance through similarity calculation and the information is encoded in the similarity matrices. The important information is then extracted by convolution and pooling operations and carried to the matching vector. Different from SCN, SAN employs an attention mechanism to capture important information in contexts. Given an utterance-response pair, SAN lets the response attend to important parts (either words or segments) in the utterance by weighting the parts using each part of the response. Each weight reflects how important the part in the utterance is regarding to the corresponding part in the response. Then for each part in the response, parts in the utterance are linearly combined with the weights, and the combination interacts with the part of the response by Hadamard product to form a representation of the utterance. Such utterance representations are computed on both a word level and a segment level. The two levels of representations are finally concatenated and processed by a GRU to form a matching vector. We theoretically analyze efficiency of SCN and SAN, and conclude that SCN is faster and easier to parallelize than SAN.

	We test the performance of SCN and SAN on two public data sets: Ubuntu Dialogue Corpus \cite{lowe2015ubuntu} and Douban Conversation Corpus \cite{wu2016sequential}. The Ubuntu corpus is a large scale English data set in which negative instances are randomly sampled and dialogues are collected from a specific domain; while the Douban corpus is a newly published Chinese data set where the conversations are crawled from an open domain forum with response candidates collected following the procedure of retrieval-based chatbots and their appropriateness judged by human annotators. Experimental results show that on both data sets, both SCN and SAN can significantly outperform the existing methods. Particularly, on the Unbuntu corpus, SCN and SAN yield $6$\% and $7$\% improvement respectively on R$_{10}$@1 over the best performing baseline method, and on the Douban corpus, the improvement on MAP from SCN and SAN over the best baseline are $2.6$\% and $3.6$\% respectively.  The results indicate that although sacrificing efficiency, SAN could be more effective than SCN in practice. Besides the quantitative evaluation, we also visualize the two models with examples from the Ubuntu corpus. The visualization reveals how the two models understand conversation contexts and provides us insights on why they can achieve big improvement over state-of-the-art methods. 

	The contributions of the work include:
	\begin{itemize}
		\item We unify the existing context-response matching models with a framework which reveals the common drawbacks they have and sheds light on our new direction.
		\item We propose a new framework for multi-turn response selection, namely sequential matching framework, which is capable of overcoming the drawbacks the existing models suffer and addressing both challenges of context-response matching in an end-to-end way.
		\item We propose a sequential convolutional network and a sequential attention network as implementations of the new framework.
		\item We conducted extensive experiments on public data sets and verified the effectiveness of the two models with both quantitative evaluation and qualitative evaluation.
		
	\end{itemize}

	The rest of the paper are organized as follows: in Section \ref{relwork} we summarize the related work and clarify the difference between this work and our previous work published on ACL conference. We formalize the learning problem in Section \ref{probform}. In Section \ref{frameold}, we interpret the existing models with a framework. Section \ref{smf} elaborates our new framework and gives two models as special cases of the framework. Section \ref{modeltrain} gives the learning objective and some training details. In Section \ref{exp} we give details of the experiments.  Finally in Section \ref{conclusion} we draw conclusions for the paper.

	\section{Related Work}\label{relwork}
	
	We briefly review the history and recent progress of chatbots, and application of text matching techniques in other tasks. Together with the review on exsiting work, we clarify the connection and difference between these work and our work in this paper.
	
	\subsection{Chatbots}
	Research on chatbots can be traced back to 1960s when ELIZA \cite{weizenbaum1966eliza}, an early chatbot, was designed with a large number of handcrafted templates and heuristic rules.  ELIZA needs huge human effort but can only return limited responses. To remedy this, researchers have developed data driven approaches \cite{higashinaka2014towards}. The idea behind data-driven approaches is to build a chatbot with the large amount of conversation data available on social media such as forums and microblogging services. Methods along this line can be categorized into retrieval based ones and generation based ones.
	
	Generation-based chatbots reply to a message with natural language generation techniques. Early work \cite{ritter2011data} regards messages and responses as source language and target language respectively, and learn a phrase-based statistical machine translation model to translate a message to a response. Recently, together with the success of deep learning approaches, the sequence-to-sequence framework has become the mainstream, because it can implicitly capture compositionality and long-span dependencies in languages. Under this framework, many models have been proposed for both single-turn conversation and multi-turn conversation. For example, in single-turn conversation, sequence-to-sequence with an attention mechanism \cite{DBLP:conf/acl/ShangLL15,vinyals2015neural} has been applied to response generation; Li et al. \cite{li2015diversity} proposed a maximum mutual information objective to improve diversity of generated responses; Xing et al. \cite{xing2016topic} and Mou et al. \cite{mou2016sequence} introduced external knowledge into the sequence-to-sequence model; Li et al. \cite{li2016persona} incorporated persona information into the sequence-to-sequence model to enhance response consistency with speakers; and Zhou et al. \cite{zhou2017emotional} explored how to generate emotional responses  with a memory augmented sequence-to-sequence model. In multi-turn conversation, Sordoni et al. \cite{sordoni2015neural} compressed a context to a vector with a multi-layer perceptron in response generation; Serban et al. \cite{serban2015building} extended the sequence-to-sequence model to a hierarchical encoder-decoder structure; and under this structure, they further proposed two variants including VHRED \cite{serban2017hierarchical} and MrRNN \cite{serban2016multiresolution} to introduce latent and explicit variables into the generation process. Upon these methods, reinforcement learning technique \cite{li2016deep} and adversarial learning technique \cite{li2017adversarial} have also been applied to response generation.

	Different from the generation based systems, retrieval-based chatbots select a proper response from an index and re-use the one to reply to  a new input. The key to response selection is how to match the input with a response. In a single-turn scenario, matching is conducted between a message and a response. For example, Hu et al. \cite{hu2014convolutional} proposed message-response matching with convolutional neural networks; Wang et al. \cite{wang2015syntax} incorporated syntax information into matching; Ji et al. \cite{ji2014information} combined a bunch of matching features, such as cosine, topic similarity, and translation score, to rank response candidates. In multi-turn conversation, matching requires taking the entire context into consideration. In this scenario, Lowe et al. \cite{lowe2015ubuntu} employed a dual LSTM model to match a response with the literal concatenation of utterances in a context; Yan et al. \cite{DBLP:conf/sigir/YanSW16} reformulated the input message with the utterances in its previous turns and performed matching with a deep neural network architecture; Zhou et al. \cite{zhou2016multi} adopted an utterance view and a word view in matching to model relationships among utterances; and Wu et al. \cite{wu2016sequential} proposed a sequential matching network that can capture important information in contexts and model relationships among utterances in a unified form.
	
	Our work belongs to retrieval based methods. It is an extension of the work \cite{wu2016sequential} published on ACL conference. In this work, we analyze the existing models from a framework view, generalize the model in \cite{wu2016sequential} to a framework, give another implementation with better performance under the framework, and compare the new model with the model in the conference paper on various aspects.
	
	\subsection{Text Matching}
	In addition to response selection in chatbots, neural network based text matching techniques have proven effective on capturing semantic relations between text pairs in a variety of NLP tasks. For example, in question answering, covolutional neural networks \cite{qiu2015convolutional,severyn2015learning} can effectively capture compositions of n-grams and their relations in questions and answers.  Inner-Attention \cite{wang2016inner} and MV-LSTM \cite{wan2015deep} can model complex interaction betwen questions and answers through recurrent neural network based architectures. More studies on text matching for question answering can be found in \cite{tan2015lstm,liu2016deep,liu2016modelling,wan2016match,he2016pairwise,yin2015abcnn,yin2015multigrancnn}. In web search, Shen et al. and Huang et al. \cite{shen2014latent,huang2013learning} built a neural network with tri-letters to alleviate mismatching of queries and documents due to spelling errors. In textual entailment, the model in \cite{rocktaschel2015reasoning} utilized a word-by-word attention mechanism to distinguish the relationship between two sentences.  Wang et al. \cite{wang2015learning} introduced another way to adopt attention mechanism for textual entailment. Besides these two work, Chen et al. \cite{chen2016enhancing}, Parikh et al. \cite{parikh2016decomposable}, and Wang et al. \cite{wang2016compare} also investigated the textual entailment problem with neural network models.
	
	In this work, we study text matching for response selection in multi-turn conversation, in which matching is conducted between a piece of text and a context which consists of multiple pieces of text dependent with each other. We propose a new matching framework which is able to extract important information in the context and model dependencies among utterances in the context.

	\section{Problem Formalization}\label{probform}
	Suppose that we have a data set $\mathcal {D} = \{(y_i,s_i,r_i)\}_{i=1}^N$, where $s_i$ is a conversation context, $r_i$ is a response candidate, and $y_i\in \{0,1\}$ is a label. $s_i=\{u_{i,1}, \ldots, u_{i,n_i}\}$ where $\{u_{i,k}\}_{k=1}^{n_i}$ are utterances. $\forall k$, $u_{i,k}=(w_{u_{i,k},1}, \ldots, w_{u_{i,k},j}, \ldots, w_{u_{i,k},n_{u_i}})$ where $w_{u_{i,k},j}$ is the $j$-th word in $u_{i,k}$ and $n_{u_i}$ is the length of $u_{i,k}$. Similarly, $r_i=(w_{r_i,1}, \ldots, w_{r_i,j},\ldots, w_{r_i,n_i})$ where $w_{r_i,j}$ is the $j$-th word in $r_i$ and $n_i$ is the length of the response. $y_i=1$ if $r_i$ is a proper response to $s_i$, otherwise $y_i=0$.  Our goal is to learn a matching model $g(\cdot,\cdot)$ with $\mathcal{D}$, and thus for any new context-response pair $(s,r)$, $g(s,r)$ measures their matching degree. According to $g(s,r)$, we can rank candidates for $s$ and select a proper one as its response.  
	
	In the following sections, we first review how the existing work defines $g(\cdot,\cdot)$ from a framework view. The framework view discloses the common drawbacks the existing work has. Then based on these analysis,  we propose a new matching framework and give two models under the framework.

	\section{A Framework for the Existing Models}\label{frameold}
	
		\begin{figure*}[t]	
			\begin{center}
				\includegraphics[width=8cm,height=6cm]{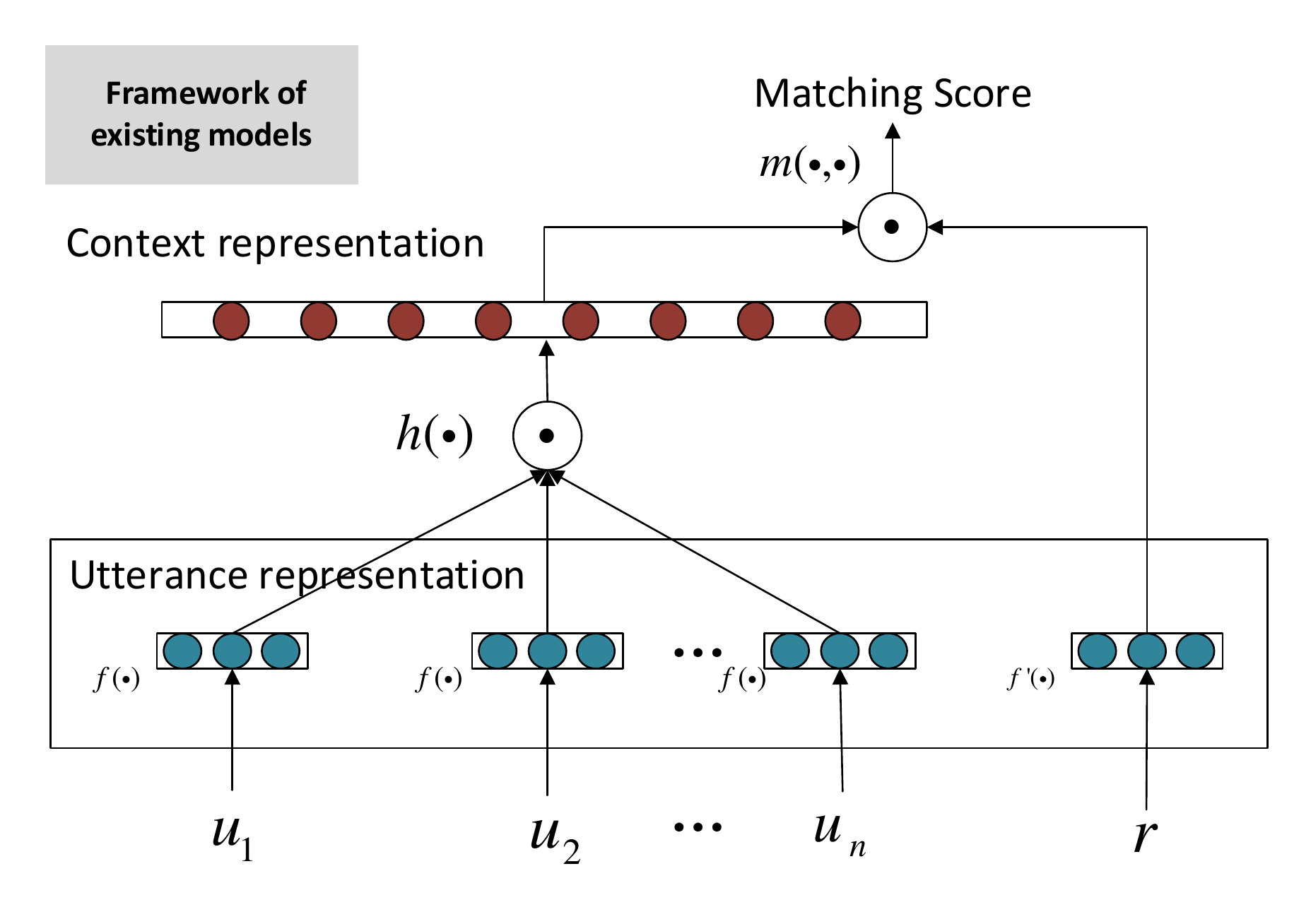}	
			\end{center}
			
			\caption{Existing models can be interpreted with a unified framework. $f(\cdot)$, $f'(\cdot)$, $h(\cdot)$, and $m(\cdot,\cdot)$ are utterance representation function, response representation function, context representation function, and matching function respectively. }\label{fig:oldarch}
		\end{figure*}
	Before us, there are a few studies on context-response matching for response selection in multi-turn conversation. For example, Lowe et al. \cite{lowe2015ubuntu} match a context and a response with recurrent neural networks (RNNs); Yan et al. \cite{DBLP:conf/sigir/YanSW16} present a deep learning to respond architecture for multi-turn response selection; and Zhou et al.  \cite{zhou2016multi} perform context-response matching from both a word view and an utterance view. Altough these models are proposed from different backgroud, we find that they can be interpreted with a unified framework given by Figure \ref{fig:oldarch}. The framework consists of utterance representation $f(\cdot)$, response representation $f'(\cdot)$, context representation $h(\cdot)$, and matching calculation $m(\cdot,\cdot)$. Given a context $s=\{u_1,\ldots, u_n\}$ and a response candidate $r$,  $f(\cdot)$ and $f'(\cdot)$ represent each $u_i$ in $s$ and $r$ as vectors or matrices by $f(u_i)$ and $f'(r)$ respectively. $\{f(u_i)\}_{i=1}^n$ are then uploaded to $h(\cdot)$ which transforms the utterance representations into $h\left(  f(u_1),\ldots, f(u_n)\right)$ as a representation of the context $s$. Finally, $m(\cdot,\cdot)$ takes $h\left(f(u_1),\ldots, f(u_n)\right)$ and $f'(r)$ as input and calculates a matching score for $s$ and $r$. To sum up, the framework performs context-response matching following a paradigm that context $s$ and response $r$ are first individually represented as vectors and then their matching degree is determined by the vectors. Under the framework, the matching model $g(s,r)$ can be defined with $f(\cdot)$, $h(\cdot)$, $f'(\cdot)$ and $m(\cdot,\cdot)$ as follows:
	\begin{equation}
	g(s,r) = m \left(h\left(f(u_1),\ldots,f(u_n)\right), f'(r)\right).
	\end{equation}

	The existing models are special cases under the framework with different definitions of $f(\cdot)$, $h(\cdot)$, $f'(\cdot)$ and $m(\cdot,\cdot)$. 
	Specifically, the RNN models in \cite{lowe2015ubuntu} can be defined as 
	\begin{equation}
	m_{rnn}(s,r) = \sigma\left(h_{rnn}\left(f_{rnn}(u_1),\ldots,f_{rnn}(u_n)\right)^\top \cdot M \cdot f'_{rnn}(r)+b\right),
	\end{equation}	
	where $M$ is a linear transformation, $b$ is a bias, and $\sigma(\cdot)$ is a sigmoid function. $\forall u_i=\{w_{u_i,1},\ldots, w_{u_i,n_i}\}$, $f_{rnn}(u_i)$ is defined by
	\begin{equation}
	f_{rnn}(u_i) = \left[\vec{w}_{u_i,1}, \ldots, \vec{w}_{u_i,k}, \ldots, \vec{w}_{u_i,n_i}\right],
	\end{equation}
	where $\vec{w}_{u_i,k}$ is the embedding of the $k$-th word $w_{u_i,k}$, and $[\cdot]$ denotes a horizontal concatenation operator on vectors or matrices\footnote{We borrow the operator from MATLAB.}. Suppose that the dimension of the word embedding is $d$, then the output of $f_{rnn}(u_i)$ is a $d\times n_i$ matrix with each column an embedding vector. Suppose that $r=(w_{r,1}, \ldots, w_{r,n_r})$, then $f'_{rnn}(r)$ is defined as
	\begin{equation}
	f'_{rnn}(r)= \text{RNN}(\vec{w}_{r,1}, \ldots, \vec{w}_{r,k}, \ldots, \vec{w}_{r,n_r}),
	\end{equation}
	where $\vec{w}_{r,k}$ is the embedding of the $k$-the word in $r$, and $\text{RNN}(\cdot)$ is either a vanilla RNN \cite{elman1990finding} or an RNN with long short-term memory (LSTM) units \cite{hochreiter1997long}. $\text{RNN}(\cdot)$ takes a sequence of vectors as an input, and outputs the last hidden state of the network. Finally, the context representation $h_{rnn} (\cdot)$ is defined by
	\begin{equation}
	h_{rnn}\left(f_{rnn}(u_1),\ldots, f_{rnn}(u_n)\right)= \text{RNN}\left([f_{rnn}(u_1), \ldots, f_{rnn}(u_n)]\right).
	\end{equation}

	In the deep learning to respond (DL2R) architecture \cite{DBLP:conf/sigir/YanSW16}, the authors first transform the context $s$ to an $s'=\{v_1,\ldots,v_o\}$ with heuristics including ``no context'', ``whole context'', ``add-one'', ``drop-out'' and ``combined''. In ``no context'', $s'=\{u_n\}$; in ``whole context'', $s'=\{u_1 \boxplus \cdots \boxplus u_n, u_n\}$ where operator $\boxplus$ glues vectors together and forms a long vector; in ``add-one'', $s'=\{u_1\boxplus u_n, \ldots, u_{n-1}\boxplus u_n, u_n\}$; in ``drop-out'', $s'=\{(c\textbackslash u_1) \boxplus u_n, \ldots, (c\textbackslash u_{n-1}) \boxplus u_{n}, u_{n}\}$ where $c=u_1\boxplus\cdots\boxplus u_n$ and $c\textbackslash u_i$ means excluding $u_i$ from $c$; and in ``combined'', $s'$ is the union of the other heuristics. Let $v_o=u_n$ in all heuristics, then the matching model of DL2R can be reformulated as 	
	\begin{equation}\label{dl2r}
	m_{dl2r}(s,r) =\sum_{i=1}^{o} \text{MLP}(f_{dl2r}(v_i) \boxplus f_{dl2r}(v_o)) \cdot \text{MLP}(f_{dl2r}(v_i) \boxplus f'_{dl2r}(r))
	\end{equation} 
	where $\text{MLP}(\cdot,\cdot)$ is a multi-layer perceptron \cite{rosenblatt1961principles}. $\forall v\in \{v_1,\ldots,v_o\}$, suppose that $\{\vec{w}_{v,1}, \ldots, \vec{w}_{v,n_v}\}$ represent embedding vectors of the words in $v$, then $f_{dl2r}(v)$ is given by
	\begin{equation}
	f_{dl2r}(v) = \text{CNN} \left(\text{Bi-LSTM} (\vec{w}_{v,1}, \ldots, \vec{w}_{v,n_v}) \right),
	\end{equation}
	where $\text{CNN}(\cdot)$ is a convolutional neural network (CNN) \cite{kim2014convolutional} and $\text{Bi-LSTM}(\cdot)$ is a bi-directional recurrent neural network with LSTM units (Bi-LSTM) \cite{graves2013speech}. The output of $\text{Bi-LSTM}(\cdot)$ is all the hidden states of the Bi-LSTM model. $f'_{dl2r}(\cdot)$ is defined in the same way with $f_{dl2r}(\cdot)$. In DL2R, $h_{dl2r}(\cdot)$ can be viewed as an identity function on $\{f_{dl2r}(v_1),\ldots, f_{dl2r}(v_o)\}$. Note that in the paper of \cite{DBLP:conf/sigir/YanSW16}, the authors also assume that each response candidate is associated with an antecedent posting $p$. This assumption does not always hold in multi-turn response selection. For example in Ubuntu Dialog Corpus \cite{lowe2015ubuntu}, there are no antecedent postings. To make the framework compatible with their assumption, we can simply extend $f_{dl2r}(r)$ to $[f_{dl2r}(p), f_{dl2r}(r)]$, and define $m_{dl2r}(s,r)$ as 
	\begin{equation}   \small
\sum_{i=1}^{o} \left(\text{MLP}(f_{dl2r}(v_i) \boxplus f_{dl2r}(v_o)) \cdot 
	 \left(\sum_p \text{MLP}(f_{dl2r}(v_i) \boxplus f_{dl2r}(p)) \cdot \text{MLP}(f_{dl2r}(v_i) \boxplus f_{dl2r}(r))\right)\right).
	\end{equation}

	Finally, in \cite{zhou2016multi}, the multi-view matching model can be re-written as
	\begin{equation} \label{multi-view}
	m_{mv}(s,r) = \sigma\left(h_{mv}(f_{mv}(u_1),\ldots, f_{mv}(u_n))^\top	\begin{bmatrix}
	M_1 \\
	M_2
	\end{bmatrix} f'_{mv}(r) + \begin{bmatrix}
	b_1 \\
	b_2
	\end{bmatrix} \right),
	\end{equation} 
	where $M_1$ and $M_2$ are linear transformations, $b_1$ and $b_2$ are biases. $\forall u_i=\{w_{u_i,1},\ldots, w_{u_i,n_i}\}$, $f_{mv}(u_i)$ is defined as
	\begin{equation}
	f_{mv}(u_i)=\{f_w(u_i), f_u(u_i)\},
	\end{equation}
	where $f_w(u_i)$ and $f_u(u_i)$ are utterance representations from a word view and an utterance view respectively. The formulation of $f_w(u_i)$ and $f_u(u_i)$ are given by 
	\begin{eqnarray*}
		f_{w}(u_i) &=& \left[\vec{w}_{u_i,1}, \ldots, \vec{w}_{u_i,n_i}\right]\\
		f_{u}(u_i)&=& \text{CNN}(\vec{w}_{u_i,1}, \ldots, \vec{w}_{u_i,n_i}).
	\end{eqnarray*}
	Suppose that $r=(w_{r,1}, \ldots, w_{r,n_r})$, then $f'_{mv}(r)$ is defined as
	\begin{equation}
	f'_{mv}(r)=[f'_w(r)^\top, f'_u(r)^\top]^\top,
	\end{equation}
	where the word view representation $f'_w(r)$ and the utterance view representation $f'_u(r)$ are formulated as
	\begin{eqnarray*}
		f'_{w}(r) &= &\text{GRU}(\vec{w}_{r,1}, \ldots, \vec{w}_{u_{r,n_r}}),\\
		f'_{u}(r) &= &\text{CNN}(\vec{w}_{r,1}, \ldots, \vec{w}_{u_{r,n_r}}),\\
	\end{eqnarray*}   
	where $\text{GRU}(\cdot)$ is a recurrent neural network with gated recurrent units \cite{cho2014learning}. The output of $f'_w(r)$ is the last hidden state of the GRU model.  The context representation $h_{mv}(f_{mv}(u_1),\ldots, f_{mv}(u_n))$ is defined as  
	\begin{equation}
	h_{mv}(f_{mv}(u_1),\ldots, f_{mv}(u_n))=[h_w(f_w(u_1),\ldots,f_w(u_n))^\top, h_u(f_u(u_1),\ldots, f_u(u_n))^\top]^\top,
	\end{equation}
	where the word view $h_w(\cdot)$ and the utterance view $h_u(\cdot)$ are defined as
	\begin{eqnarray*}
		h_{w}(f_w(u_1),\ldots,f_w(u_n))&=&\text{GRU}\left([f_{w}(u_1), \ldots, f_{w}(u_{n})]\right),\\
		h_{u}(f_u(u_1),\ldots, f_u(u_n))&=&\text{GRU}\left(f_{u}(u_1), \ldots, f_{u}(u_{n})\right).
	\end{eqnarray*}  
	
	There are several advantages when applying the framework view to the existing context-response matching models. First, it unifies the existing models and reveals the instinct connections among them. These models are nothing but similarity functions of a context representation and a response representation. Their difference on performance comes from how well the two representations capture the semantics and the structures of the context and the response and how accurate the similarity calcuation is. For example, in empirical studies, the multi-view model performs much better than the RNN models. This is because the multi-view model captures the sequential relationship among words, the composition of n-grams, and the sequential relationship of utterances by $h_w(\cdot)$ and $h_u(\cdot)$; while in RNN models, only the sequential relationship among words are modeled by $h_{rnn}(\cdot)$.  Second, it is easy to make an extension of the existing models by replacing $f(\cdot)$, $f'(\cdot)$, $h(\cdot)$, and $m(\cdot,\cdot)$. For example, we can replace the $h_{rnn}(\cdot)$ in RNN models with a composition of CNN and RNN to model both composition of n-grams and their sequential relationship, and we can replace the $m_{rnn}(\cdot)$ with a more powerful neural tensor network \cite{socher2013reasoning}. Third, the framework unveils the limitations the existing models and their possible extentions suffer: everything in the context are compressed to one or more fixed-length vectors before matching; and there is no interaction between the context and the response in the formation of their representations. The context is represented without enough supervision from the response, and so is the response.  As a result, these models may lose important inforamtion of contexts in matching, and more seriously, no matter how we improve them, as long as the improvement is under the framework, we cannot overcome the limitations. The framework view motivates us to propose a new framework that can essentially change the existing matching paradigm.

	\section{Sequential Matching Framework}\label{smf}
	\begin{figure*}[t]	
		\begin{center}
			\includegraphics[width=8cm,height=6cm]{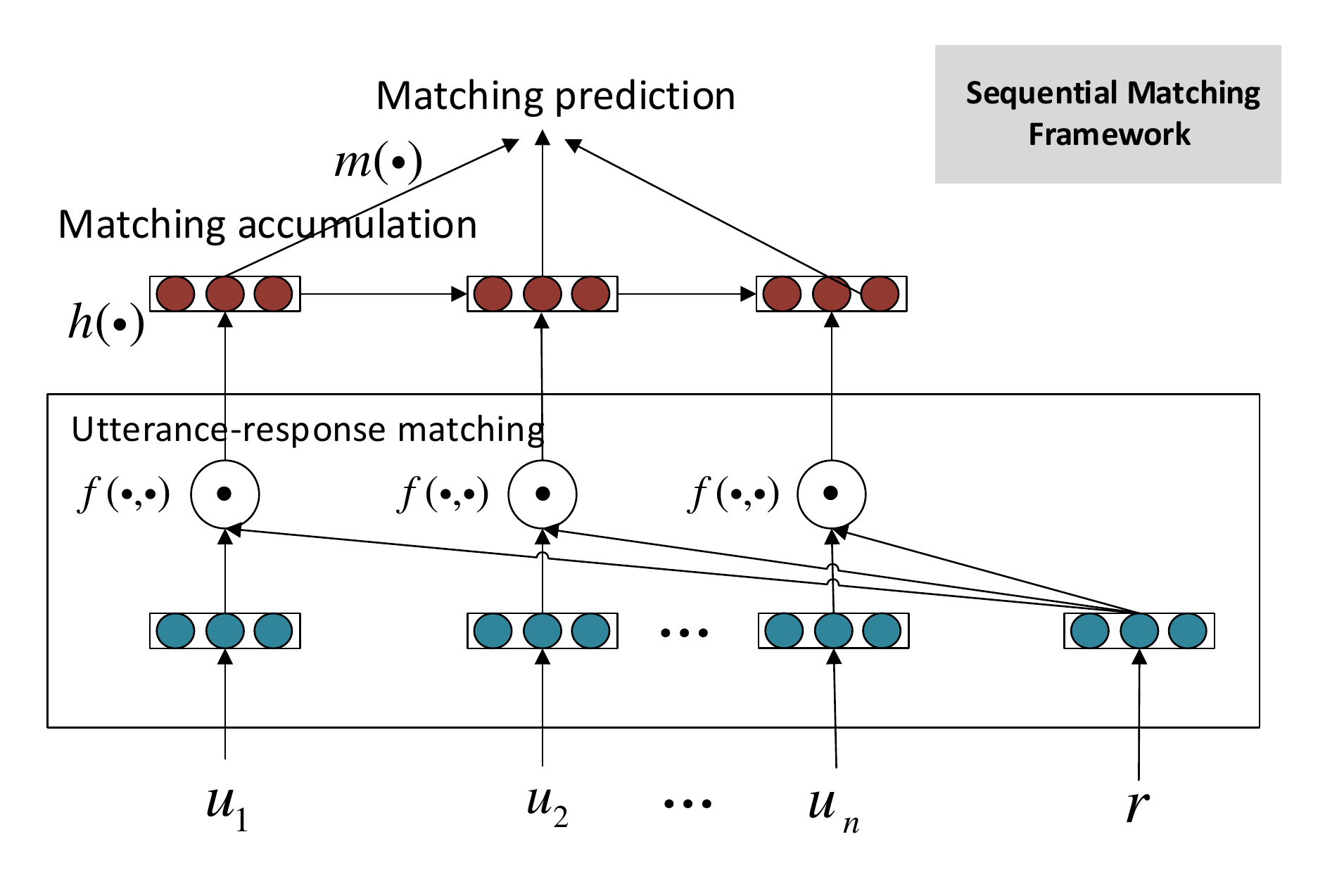}	
		\end{center}
		
		\caption{Our new framework for multi-turn response selection, which is called Sequential Matching Framework. It first computes a matching vector between an utterance and a response, then the matching vectors are accumulated by a GRU. Finally, the matching score is obtained with the hidden states in the second layer. }\label{fig:newarch}
	\end{figure*}
	We propose a sequential matching framework (SMF) that can simultaneously capture important information in a context and model rationships among utterances in the context. Figure \ref{fig:newarch} gives the architecture of SMF. SMF consists of utterance-response matching $f(\cdot,\cdot)$, matching accumulation $h(\cdot)$, and matching prediction $m(\cdot)$. The three components are organized in a three-layer architecture. Given a context $s=\{u_1,\ldots, u_n\}$ and a response candidate $r$, the first layer matches each $u_i$ in $s$ with $r$ through $f(\cdot,\cdot)$ and forms a sequence of matching vectors $\{f(u_1,r), \ldots, f(u_n,r)\}$. Here, we require $f(\cdot,\cdot)$ to be capable of differentiating important parts from unimportant parts in $u_i$ and carry the important information into $f(u_i, r)$. Details of how to design such a $f(\cdot, \cdot)$ will be described later. The matching vectors $\{f(u_1,r), \ldots, f(u_n,r)\}$ are then uploaded to the second layer where $h(\cdot)$ models relationships and dependencies among the utterances $\{u_1,\ldots u_n\}$. Here, we define $h(\cdot)$ as a recurrent neural network whose output is a sequence of hidden states $\{h_1, \ldots, h_n\}$. $\forall  k \in \{1,\ldots, n\}$, $h_k$ is given by
	\begin{equation}\label{matchacu}
	h_k =h'\bigg(h_{k-1}, f(u_{k},r) \bigg),
	\end{equation}
	where $h'(\cdot, \cdot)$ is a non-linear transformation, and $h_0 = 0$. $h(\cdot)$ accumulates matching vectors $\{f(u_1,r), \ldots, f(u_n,r)\}$ in its hidden states. Finally, in the third layer, $m(\cdot)$ takes $\{h_1, \ldots, h_n\}$ as an input and predicts a matching score for $(s,r)$. In brief, SMF matches $s$ and $r$ with a $g(s,r)$ defined as
	\begin{equation}\label{newframe}
	g(s,r) = m \bigg ( h \Big (  f(u_{1},r),f(u_{2},r),\ldots, f(u_{n_i}，r) \Big ) \bigg).
	\end{equation}
	
	SMF makes two major differences over the existing framework: first, SMF lets each utterance in the context and the response ``meet'' at the very beginning, and therefore, utterances and the response can sufficiently interact with each other. Through the interaction, the response will help recognize important information in each utterance. The information is preserved in the matching vectors and carried into the final matching score with minimal loss; second, matching and utterance relationships are coupled rather than separately modeled as in the existing framework. Hence, the utterance relathinships (e.g., the order of the utterances), as a kind of knowledge, can supervise the formation of the matching score. Because of the differences, SMF can overcome the drawbacks the existing models suffer and tackle the two challenges of context-response matching simultaneously. 
	
	It is obvious that  the success of SMF lies in how to design $f(\cdot,\cdot)$, because $f(\cdot,\cdot)$ plays a key role in capturing important information in a context. In the following secions, we will first specify the design of $f(\cdot,\cdot)$, and then discuss how to define $h(\cdot)$ and $m(\cdot)$.

	\subsection{Utterance-Response Matching}\label{multi-channel}
	We design the utterance-response matching fucntion $f(\cdot, \cdot)$ in SMF as neural networks to benefit from their powerful represenation abilities. To guarantee that $f(\cdot,\cdot)$ can capture important information in utterances with the help of the response, we implement $f(\cdot,\cdot)$ using a convolution-pooling technique and an attention technique, which results in a sequential convolutional network (SCN) and a sequential attention network (SAN). Moerover, in both SCN and SAN, we consider matching on multiple levels of granulatiry of text.     
	
	\begin{figure*}[t]	
		\begin{center}
			\includegraphics[width=14cm,height=5.5cm]{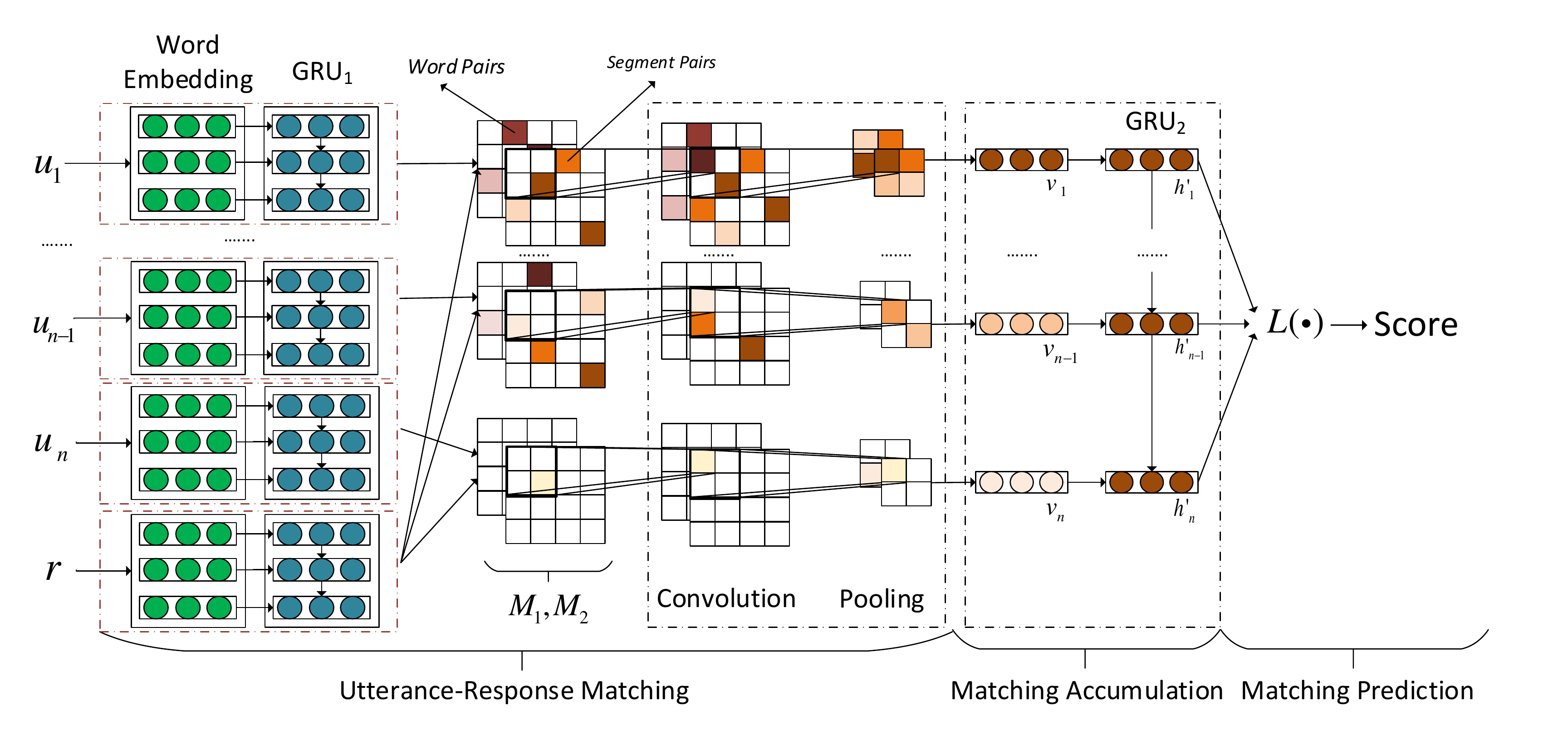}	
		\end{center}
		
		\caption{The architecture of SCN. The first layer extracts matching information from interactions between utterances and a response on a word level and a segment level by a CNN. The second layer accumulates the matching information from the first layer by a GRU. The third layer takes the hidden states of the second layer as an input and calculates a matching score. }\label{fig:scn}
	\end{figure*}
	\subsubsection{Sequential Convolutional Network}
	Figure \ref{fig:scn} gives the architecture of SCN. Given an utterance $u$ in a context $s$ and a response candidate $r$, SCN looks up an embedding table and represents $u$ and $r$ as $\mathbf{U} = \left[e_{u,1},\ldots,e_{u,n_u}\right]$ and $\mathbf{R} = \left[e_{r,1},\ldots,e_{r,n_r}\right]$ respectively, where $e_{u,i}, e_{r,i} \in \mathbb{R} ^ d$ are the embeddings of the $i$-th word of $u$ and $r$ respectively. With $\mathbf{U}$ and $\mathbf{R}$, SCN constructs a word-word similarity matrix $\mathbf{M}_1 \in \mathbb{R}^{n_u \times n_r}$ and a sequence-sequence similarity matrix $\mathbf{M}_2$ $ \in \mathbb{R}^{n_u \times n_r}$ as two input channels of a convolutional neural network (CNN). The CNN then extracts important matching information from the two matrices and encodes the information into a matching vector $v$.
	
	Specifically, $\forall i,j$, the $(i,j)$-th element of $\mathbf{M}_1$ is defined by
	\begin{equation}\label{M1Element}
	e_{1,i,j} = e_{u,i}^{\top} \cdot e_{r,j}. 
	\end{equation}
	$\mathbf{M}_1$ models the interaction between $u$ and $r$ on a word level. 
	
	To get $\mathbf{M}_2$, we first transform $\mathbf{U}$ and $\mathbf{R}$ to sequences of hidden vectors with a GRU. Suppose that $\mathbf{H}_u = \left[h_{u,1},\ldots, h_{u,n_u}\right]$ are the hidden vectors of $\mathbf{U}$, then $\forall i$, $h_{u,i}\in \mathbb{R}^m$ is defined by
	
	\small	 
	\begin{eqnarray}\label{gru}
	&& z_i = \sigma(\mathbf{W_z} e_{u,i} + \mathbf{U_z} {h}_{u,i-1}) \nonumber \\
	&& r_i = \sigma(\mathbf{W_r} e_{u,i} + \mathbf{U_r} {h}_{u,i-1}) \nonumber \\
	&&\widetilde{h}_{u,i} = tanh(\mathbf{W_h} e_{u,i} + \mathbf{U_h} (r_i \odot {h}_{u,i-1}))\nonumber\\
	&& h_{u,i} = z_i \odot \widetilde{h}_{u,i} + (1-z_i) \odot {h}_{u,i-1},
	\end{eqnarray}
	\normalsize	 	 
	where $h_{u,0}=0$, $z_i$ and $r_i$ are an update gate and a reset gate respectively, $\sigma(\cdot)$ is a sigmoid function, and $\mathbf{W_z}$, $\mathbf{W_h}$, $\mathbf{W_r}$, $\mathbf{U_z}$, $\mathbf{U_r}$,$ \mathbf{U_h}$ are parameters. Similarly, we have $\mathbf{H}_r = \left[h_{r,1},\ldots, h_{r,n_r}\right]$ as the hidden vectors of $\mathbf{R}$. 
	Then, $\forall i,j$,  the $(i,j)$-th element of $\mathbf{M}_2$ is defined by
	\begin{equation}\label{M2Element}
	\small
	e_{2,i,j} = h_{u,i}^{\top} \mathbf{A}  h_{r,j},
	\end{equation}	 
	where $\mathbf{A} \in \mathbb{R}^{m \times m}$ is a linear transformation. $\forall i$, GRU encodes the sequential information and the dependency among words until position $i$ in $u$ into the $i$-th hidden state. As a consequence, $\mathbf{M}_2$ models the interaction between $u$ and $r$ on a segment level.

	$\mathbf{M}_1$ and $\mathbf{M}_2$ are then processed by a CNN to compute the matching vector $v$. $\forall f=1,2$, CNN regards $\mathbf{M}_f$ as an input channel, and alternates convolution and max-pooling operations. Suppose that $z^{(l,f)} = \left[ z^{(l,f)}_{i,j}  \right]_{I^{(l,f)} \times J^{(l,f)}}$ denotes the output of feature maps of type-$f$ on layer-$l$, where $z^{(0,f)}= \mathbf{M}_f$, $\forall f = 1,2$. On the convolution layer, we employ a 2D convolution operation with a window size ${r_w^{(l,f)} \times r_h^{(l,f)}}$, and define $z_{i,j}^{(l,f)}$ as
	\begin{equation}\label{Conv}
	z_{i,j}^{(l,f)} = \sigma (\sum_{f'=0}^{F_{l-1}} \sum_{s=0}^{r_w^{(l,f)}} \sum_{t=0}^{r_h^{(l,f)}} \mathbf{W}_{s,t} ^ {(l,f)} \cdot z_{i+s,j+t} ^{(l-1,f')} + \mathbf{b}^{l,k} ), 
	\end{equation}
	where $\sigma(\cdot)$ is a ReLU, $\mathbf{W} ^ {(l,f)} \in \mathbb{R}^{r_w^{(l,f)} \times r_h^{(l,f)}} $ and $\mathbf{b}^{l,k}$ are parameters, and $F_{l-1}$ is the number of feature maps on the $(l-1)$-th layer. A max pooling operation follows a convolution operation and can be formulated as 
	\begin{equation}\label{Pool}
	z_{i,j}^{(l,f)} = \max_{p_w^{(l,f)} >  s \geq 0} \max_{p_h^{(l,f)} > t \geq 0} z_{i+s,j+t} , 
	\end{equation}
	where $p_w^{(l,f)}$ and $p_h^{(l,f)}$ are the width and the height of the 2D pooling respectively. The matching vector $v$ is defined by concatenating outputs of the last feature maps and transforming it to a low dimensional space:
	\begin{equation}\label{transform}
	v = \mathbf{W_c}[z_{0,0}^{l',1} \ldots z_{I,J}^{l',1},z_{0,0}^{l',2} \ldots z_{I,J}^{l',2}] + \mathbf{b_c} , 
	\end{equation}
	where $l'$ denotes the last layer, and $I$ and $J$ are the maximum indices of the feature map. $\mathbf{W_c}$ and $\mathbf{b_c}$ are parameters.

	SCN distills important information in each utterance in the context from multiple levels of granularity through convolution and pooling operations on similarity matrices.  From Equation (\ref{M1Element}), (\ref{M2Element}), (\ref{Conv}), and (\ref{Pool}), we can see that by learning word embeddings and parameters of GRU from training data, important words or segments in the utterance may have high similarity with some words or segments in the response and result in high value areas in the similarity matrices. These areas will be transformed and extracted to the matching vector by convolutions and poolings. We will further explore the mechanism of SCN by visualizing $\mathbf{M}_1$ and $\mathbf{M}_2$ of an example in Section \ref{exp}. 
	
	\begin{figure*}[h]	
		\begin{center}
			\includegraphics[width=12cm,height=6cm]{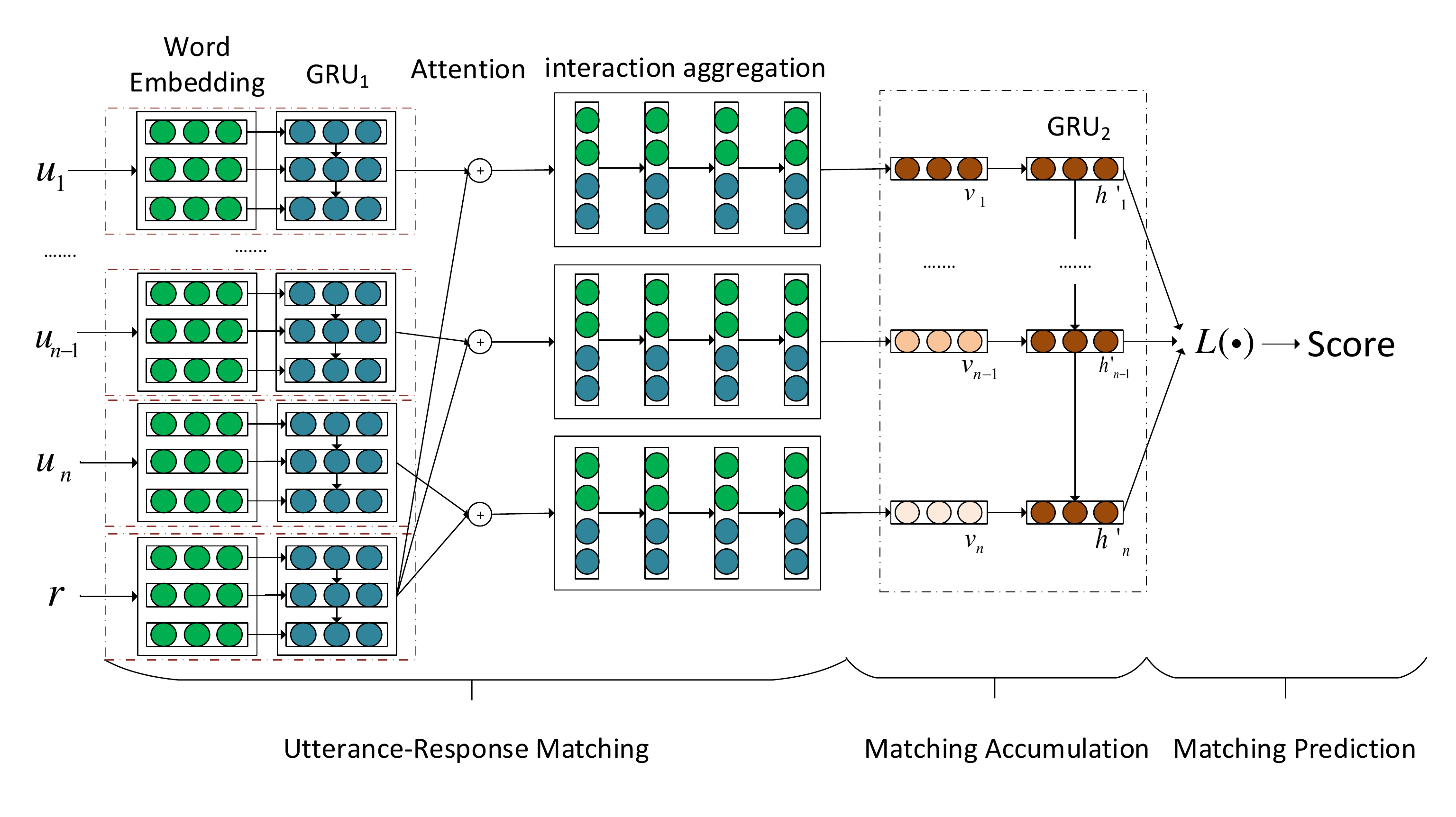}	
		\end{center}

		\caption{The architecture of SAN. The first layer highlights important words and segments in context, and computes a matching vector from both of word level and segment level. Similar to SCN, the second layer employs a GRU to accumulate the matching information, and the third layer predicts the final matching score.}\label{fig:san}
	\end{figure*}
	\subsubsection{Sequential Attention Network}  
	With word embeddings $\mathbf{U}$ and $\mathbf{R}$ and hidden vectors $\mathbf{H}_u$ and $\mathbf{H}_r$, SAN also performs utterance-response matching on a word level and a segment level. Figure \ref{fig:san} gives the architecture of SAN. In each level of matching, SAN exploits every part of the response (either a word or a hidden state) to weight the parts of the utterance and obtain a weighted representation of the utterance. The utterance representation then interacts with the part of the response. The interactions are finally aggregated following the order of the parts in the response as a matching vector. 
	
	Specifically, $\forall  e_{r,i} \in \mathbf{R}$, the weight of  $e_{u,j} \in \mathbf{U}$ is given by
	\begin{eqnarray} \label{word_attention}
	&& \omega_{i,j} = v^{\top} tanh(e_{u,j}^{\top}\mathbf{ W_{att1}} e_{r,i} + \mathbf{b_{att1}})\\
	&& \alpha_{i,j} = \frac{e^{\omega_{i,j}}}{\sum_{j=1}^{n_u} e^{\omega_{i,j}}}  ,
	\end{eqnarray}
	where $\mathbf{ W_{att1}} \in \mathbb{R}^{d \times d} $, $v \in \mathbb{R}^d$, and $ \mathbf{b_{att1}} \in \mathbb{R}^d$ are parameters. $\omega_{i,j} \in \mathbb{R}$ represents the importance of $e_{u,j}$ in the utterance corresponding to $e_{r,i}$ in the response. $\alpha_{i,j}$ is normalized importance. The interaction between $u$ and $e_{r,i}$ is then defined as
	\begin{equation} \label{pair-wise vector}
	t_{1,i} =  \left(\sum_{j=1}^{n_u} \alpha_{i,j} {e_{u,j}}\right) \odot e_{r,i},
	\end{equation}
	where $(\sum_{j=1}^{n_u} \alpha_{i,j} {e_{u,j}})$ is the representation of $u$ with weights $\{\alpha_{i,j}\}_{j=1}^{n_u}$, and $\odot$ is Hadamard product. 
	
	Similarly, $\forall  h_{r,i} \in \mathbf{H}_r $, the weight of  $h_{u,j} \in \mathbf{H}_u$ can be defined as
	\begin{eqnarray} \label{hidden_attention}
	&& \omega'_{i,j} = v'^{\top} tanh(h_{u,j}^{\top}\mathbf{ W_{att2}} h_{r,i} + \mathbf{b_{att2}})\\
	&& \alpha'_{i,j} =\frac{e^{\omega'_{i,j}}}{\sum_{j=1}^{n_u} e^{\omega'_{i,j}}}  ,
	\end{eqnarray}
	where $\mathbf{W_{att2}} \in \mathbb{R}^{d \times d} $, $v' \in \mathbb{R}^d$, and $ \mathbf{b_{att2}} \in \mathbb{R}^d$ are parameters. The interaction between $u$ and $h_{r,i}$ then can be formulated as 
	\begin{equation}
	t_{2,i} =  \sum_{j=1}^{n_u} \left(\alpha'_{i,j} {h_{u,j}}\right) \odot h_{r,i}.
	\end{equation}
	We denote the attention weights $\{\alpha_{i,j} \}$ and $\{\alpha'_{i,j} \}$ as $\mathbf{A_1}$  and $\mathbf{A_2}$ respectively. 
	With the word-level interaction $\mathbf{T}_1 = [t_{1,1}, \ldots, t_{1,n_r}]$ and the segment level interaction $\mathbf{T}_2 = [t_{2,1}, \ldots, t_{2,n_r}]$, we form a $\mathbf{T} = [t_1, \ldots, t_{n_r}]$ by defining $t_i$ as $[t_{1,i}^\top,t_{2,i}^\top]^\top$. The matching vector $v$ of SAN is then obtained by processing $\mathbf{T}$ with a GRU:
	\begin{equation} \label{attention}
	v = \text{GRU}(\textbf{T}),
	\end{equation}
	where the specific parameterization of $\text{GRU}(\cdot)$ is similar to Equation (\ref{gru}), and we take the last hidden state of the GRU as $v$. 
	
	From Equation (\ref{word_attention}) and Equation (\ref{hidden_attention}), we can see that SAN identifies important information in utterances in a context through an attention mechanism. Words or segments in utterances that are useful to recognize the appropriateness between the context and a response will receive high weights from the response. The information conveyed by these words and segments will be highlighted in the interaction between the utterances and the response and carried to the matching vector through a recurrent neural network which models the aggregation of information in the utterances under the supervision of the response.  Similar to SCN, we will further investigate the effect of the attention mechanism in SAN by visualizing the attention weights in Section \ref{exp}.

	\subsubsection{SAN v.s. SCN} \label{section:vs}
	Since SCN and SAN exploits different mechanisms to understand important parts in contexts, an interesting question arises: what the advandages and disadvantages of the two models are in practice. Here, we leave empirical comparison of their performance to experiments and first compare SCN with SAN on the follow aspects: (1) amount of parallelable computation which is measured by the miminum number of sequential operations requried; and (2) total time complexity.  
	
	Table \ref{table:comp} summarizes the comparsion between the two models. In terms of parallelability, SAN uses two RNNs to learn the representations which requires $2n$ sequential operations, whereas SCN has $n$ sequentially executed operations in the construction of $\mathbf{M}_2$. Hence, SCN is easier to parallelize than SAN. In terms of time complexity, the complexity of SCN is $\mathcal{O}(k \cdot n \cdot d^2 + n \cdot d^2 + n^2 \cdot d)$, where $k$ is the number of feature maps in convolutions, $n$ is $max(n_u,n_r)$, and $d$ is embedding size. More specifically, in SCN, the cost on construction of $\mathbf{M}_1$ and $\mathbf{M}_2$ is $\mathcal{O} (n \cdot d^2 + n^2 \cdot d)$, and the cost on convolution and pooling is $\mathcal{O}(k \cdot n \cdot d^2)$. The complexity of SAN is  $\mathcal{O}(n^2 \cdot d +n^2 \cdot d^2)$, where $\mathcal{O}(n^2 \cdot d)$ is the cost on calculating $\mathbf{H}_u$ and $\mathbf{H}_r$ and $\mathcal{O}(n^2 \cdot d^2)$ is the cost of the following attention based GRU. In practice, $k$ is usually much smaller than the maximum sentence length $n$. Therefore, SCN could be faster than SAN. The conclusion is also verified by empircal results in Section \ref{exp}. 

	\begin{table}[h]
		\caption{Comparison between SCN and SAN. $k$ is the kernel number of convolutions. $n$ is $max(n_u,n_r)$. $d$ is the embedding size.} \label{table:comp} 
			\begin{tabular}{c|c|c} \hline
				& time complexity & number of sequential operations \\\hline
				SCN & $\mathcal{O}(k \cdot n \cdot d^2 + n \cdot d^2 + n^2 \cdot d)$ &  $ n $  \\ \hline
				SAN & $\mathcal{O}(n^2 \cdot d^2 + n^2 \cdot d )$ &   $2n$  \\ \hline
		\end{tabular}
	
	\end{table}

	\subsection{Matching Accumulation}
	The function of matching accumulation $h(\cdot)$ in SMF can be implemented with any recurrent neural networks such as LSTM and GRU. In this work, we fix $h(\cdot)$ as GRU in both SCN and SAN. Given $\{f(u_1,r), \ldots, f(u_n,r)\}$ as the output of the first layer of SMF, the non-linear transformation $h'(\cdot, \cdot)$ in Equation (\ref{matchacu}) is formulated as
	\begin{eqnarray}\label{gru2}
	&& z_i' = \sigma(\mathbf{W_z}' f(u_i,r) + \mathbf{U_z}' {h}_{i-1}) \nonumber \\
	&& r_i' = \sigma(\mathbf{W_r}' f(u_i,r) + \mathbf{U_r}' {h}_{i-1}) \nonumber \\
	&&\widetilde{h}_{i} = tanh(\mathbf{W_h}' f(u_i,r) + \mathbf{U_h}' (r_i \odot {h}'_{i-1}))\nonumber\\
	&& h_{i} = z_i \odot \widetilde{h}_{i} + (1-z_i) \odot {h}_{i-1},
	\end{eqnarray}
	where $\mathbf{W_z}'$, $\mathbf{W_h}'$, $\mathbf{W_r}'$, $\mathbf{U_z}'$, $\mathbf{U_r}'$,$ \mathbf{U_h}'$ are parameters, and $z_i'$ and $r_i'$ are an update gate and a reset gate respectively. Here, $h_i$ is a hidden state, which encodes the matching information in its previous turns. From Equation (\ref{gru2}), we can see that the reset gate (i.e., $r_i$) and the update gate (i.e., $z_i$) control how much information from the current matching vector $f(u_i,r)$ flows into the accumulation vector $h_i$. Ideally, the two gates should let matching vectors that correspond to important utterances make much impact to the accumulation vectors (i.e., the hidden states) while block the information from the unimportant utterances.  In practice, we find that we can achieve this by learning SCN and SAN from large scale of conversation data. The details will be given in Section \ref{exp}.

	\subsection{Matching Prediction}\label{sec:matching}
	$m(\cdot)$ takes $\{h_1, \ldots, h_n\}$ from $h(\cdot)$ as an input and predicts a matching score for $(s,r)$. We consider three approaches to implementing $m(\cdot)$. 
	
	\subsubsection{Last State} The first approach is that we only use the last hidden state $h_n$ to calculate a matching score. The underlying assumption is that important information in the context, after selected by the gates of the GRU, has been encoded into the vector $h_n$. Then $m(\cdot)$ is formulated as: 
	\begin{equation}
	m_{last}(h_1, \ldots, h_n) = softmax (\mathbf{W_l} h_n + \mathbf{b_l}),
	\end{equation}
	where $\mathbf{W_l}$ and $\mathbf{b_l}$ are parameters. 
	
	\subsubsection{Static Average} The second approach is combining all hidden states with weights determined by their positions. In this approach, $m(\cdot)$ can be formulated as
	\begin{equation}
	m_{static}(h_1,\ldots,h_n) = softmax (\mathbf{W_s}(\sum_{i=1}^n w_i h_i)+ \mathbf{b_s}),
	\end{equation}
	where $\mathbf{W_s}$ and $\mathbf{b_s}$ are parameters, and $w_i$ is the weight of the $i$-th hidden state and learnt from data. Note that in $m_{static}(\cdot)$, once $\{w_i\}_{i=1}^n$ are learnt, they are fixed for any $(s,r)$ pairs, and that is why we call the approach ``static average''. Compared to last state, static average can leverage more information in the early parts of  $\{h_1, \ldots, h_n\}$, and thus can avoide information loss from the process of the GRU in $h(\cdot)$. 
	
	\subsubsection{Dynamic Average} Similar to static average, we also combine all hidden states to calculate a matching score, but the difference is that the combination weights are dynamcially computed by the hidden states and the utterance vectors through an attention mechansim as in \cite{bahdanau2014neural}. The weights will change according to the content of the utterances in different contexts, and that is why we call the approach ``dynamic average''. In this approach, $m(\cdot)$ is defined as   
	\begin{eqnarray} \label{dynamic_average}
	&& t_i =t_s^{\top} tanh(\mathbf{W_{d1}} h_{u,n_u} + \mathbf{W_{d2}} h_i + \mathbf{b_{d1}}),\nonumber \\
	&& \alpha_i = \frac{exp(t_i )}{\sum_i exp(t_i)},\nonumber \\
	&& m(h_1,\ldots, h_n) = softmax (\mathbf{W_d}(\sum_{i=1}^n \alpha_i h_i)+ \mathbf{b_{d2}}),
	\end{eqnarray}
	where $\mathbf{W_{d1}} \in \mathbb{R}^{q \times m}, \mathbf{W_{d2}} \in \mathbb{R}^{q \times q}$, $\mathbf{b_{d1}} \in \mathbb{R}^q$, $\mathbf{W_d}  \in \mathbb{R}^{q \times q}$, and $\mathbf{b_{d2}} \in \mathbb{R}^{q }$  are parameters. $t_s$ is a virtual context vector which is learned in training. $h_i$ and $h_{u,n_u}$ are $i$-th hidden state of $h(\cdot)$ and the final hidden state of the utterance respectively. 
	

	\section{Model Training} \label{modeltrain}
	We choose cross entropy as the loss function. Let $\Theta$ denote the parameters of $f(\cdot,\cdot)$, $h(\cdot,\cdot)$ and $m(\cdot)$, then the objective function  $\mathcal{L}(\mathcal{D},\Theta)$ can be written as
	
	\begin{equation}\label{obj}
	\mathcal{L}(\mathcal{D},\Theta) =	- \sum_{i=1}^{N} \left[y_i log(g(s_i,r_i)) + (1-y_i)log(1-g(s_i,r_i))\right],
	\end{equation}
	where $N$ in the number of instances in $\mathcal{D}$. We optimize the objective function using back-propagation and the parameters are updated by stochastic gradient descent with Adam algorithm \cite{kingma2014adam}.  The parameters are updated by stochastic gradient descent with Adam algorithm \cite{kingma2014adam} on a single Tesla K80 GPU. The initial learning rate is $0.001$, and the parameters of Adam, $\beta_1$ and $\beta_2$ are $0.9$ and $0.999$ respectively. We employ early-stopping as a regularization strategy. Models are trained in mini-batches with a batch size of $200$. 
	
	\section{Experiments} \label{exp}
	We test SAN and SCN on two public data sets with both quantitative metrics and qualitative analysis. 
	\subsection{Data Sets}
The first data set we exploited to test the performance of our models is Ubuntu Dialogue Corpus \cite{lowe2015ubuntu}. The corpus contains large scale two-way conversations collected from the chat logs of Ubuntu forum. The conversations are multi-turn
discussions about Ubuntu-related technical issues. We used the copy shared by Xu et al. \cite{xu2016incorporating} \footnote{\url{https://www.dropbox.com/s/2fdn26rj6h9bpvl/ubuntu data.zip?dl=0}}, in which numbers, urls, and paths are replaced by special placeholders. The data set consists of $1$ million context-response pairs for training, $0.5$ million pairs for validation, and $0.5$ million pairs for test. In each conversation, a human reply is selected as a positive response to the context, and negative responses are randomly sampled. The ratio of postive responses and negative responses is $1:1$ in the training set, and $1:9$ in both the validation set and the test set.

In addition to Ubuntu Dialogue Corpus, we selected Douban Conversation Corpus \cite{wu2016sequential} as another data set. The data is a recently released large scale open domain conversation corpus in which conversations are crawled from a popular Chinese forum Douban Group \footnote{\url{https://www.douban.com/group/}}. The training set contains $1$ million context-response pairs, and the validation set contains $5$ thousand pairs. In both sets, a context has a human reply as a postive response and a randomly sampled reply as a negative response. Therefore, the ratio of postive instances and negative instances in both training and validation is $1:1$. Different from Ubuntu Dialogue Corpus, the test set of Douban Conversation Corpus contains $1,000$ contexts with each one having $10$ responses retrieved from an pre-built index. Each response receives three labels from human annotators which indicate its appropriateness as a reply to the context and the majority of the labels is taken as the final decision. The Fleiss' kappa \cite{fleiss1971measuring} of the labeling is $0.41$, which means that the labelers reached a relatively high agreement in their work. Note that in our experiments, we removed contexts whose responses are all labeled as positive or negative. After this step, there are $6,670$ context-response pairs left in the test set.

Table \ref{dataset} summarizes the statistics of the two data sets.

	\begin{table*}
		\caption{Statistics of the two data sets }\label{dataset}
		\centering
		\begin{tabular}{c|c|c|c|c|c|c}
			\thickhline
			& \multicolumn{3}{c|}{\textbf{Ubuntu Corpus}}    &        \multicolumn{3}{c}{\textbf{Douban Corpus}}        \\ \hline
			&train&val&test &train&val&test\\ \hline
			$\#$ context-response pairs &1M& 0.5M & 0.5M &1M&50k & 10k\\ \hline
			$\#$  candidates per context & 2&10&10 & 2&2&10\\ \hline
			$\#$  positive candidates per context&1&1&1  &1&1& 1.18\\ \hline
			Min. $\#$  turns per context  & 3&3&3 & 3&3&3\\ \hline
			Max. $\# $  turns per context  &19 & 19 & 19 & 98&91&45\\\hline
			Avg. $\#$ turns per context  & 10.10&10.10& 10.11 & 6.69&6.75&6.45\\\hline
			Avg. $\#$  words per utterance & 12.45&12.44 & 12.48 &18.56&18.50 & 20.74\\ \hline
			
		\end{tabular}					
	\end{table*}

	\subsection{Baselines}
	
	We compared our methods with the following methods:

	\textbf{TF-IDF}: we followed Lowe et al. \cite{lowe2015ubuntu} and computed tf-idf based cosine similarity between a context and a response. Utterances in the context are concatenated to form a document. Idf is computed on the training data. 
	
	\textbf{Basic deep learning models}: we employed models in \cite{lowe2015ubuntu} and \cite{kadlec2015improved}, in which representations of a context are learnt by neural networks with the concatenation of utterances as inputs and the final matching score is computed by a bilinear function of the context representation and the response representation. Models including RNN, CNN, LSTM and BiLSTM were selected as baselines. 
	
	\textbf{Multi-View}: the model proposed by Zhou et al. \cite{zhou2016multi} that utilizes a hierarchical recurrent neural network to model utterance relationships. It integrates information in a context from an utterance view and a word view. Details of the model can be found in Equation (\ref{multi-view}).  
	
	\textbf{Deep learning to respond (DL2R)}: the authors in \cite{DBLP:conf/sigir/YanSW16} proposed several approaches to reformulate a message with previous turns in a context. The response and the reformulated message are then represented by a composition of RNN and CNN. Finally, the matching score is computed with the concatenation of the representations. Details of the model can be found in Equation (\ref{dl2r})
	
	\textbf{Advanced single-turn matching models}: since  BiLSTM does not represent the state-of-the-art matching model, we concatenated the utterances in a context and matched the long text with a response candidate using more powerful models including MV-LSTM \cite{wan2016match} (2D matching), Match-LSTM \cite{wang2015learning}, Attentive-LSTM \cite{tan2015lstm} (two attention based models). To demonstrate the importance of modeling utterance relationships, we also calculated a matching score for the concatenation of utterances and the response candidate using the methods in Section  \ref{multi-channel}. The two models are simple versions of SCN and SAN respectively without considering utterance relationships. We denote them as  SCN$_{single}$ and SAN$_{single}$ respectively.

    \subsection{Evaluation Metrics}
	In experiments on the Ubuntu corpus, we followed \cite{lowe2015ubuntu} and used recall at position $k$ in $n$ candidates ($R_n@k$) as evaluation metrics. Here the matching models are required to return $k$ most likely responses, and $R_n@k = 1$ if the true response is among the $k$ candidates. $R_n@k$ will become larger when $k$ gets larger or $n$ gets smaller. 
	
	On the Douban corpus, apart from $R_n@k$s, we also followed the convention of information retrieval and used mean average precision (MAP) \cite{baeza1999modern}, mean reciprocal rank (MRR) \cite{voorhees1999trec}, and precision at position 1 (P@1) as evaluation metrics. We did not calculate $R_2@1$ on the test data because in Douban corpus one context could have more than one correct responses, and we have to randomly sample one for $R_2@1$, which may bring bias to evaluation. 
	
	\subsection{Parameter Tuning} \label{parameter}
	For baseline models, we copied the numbers in the existing papers if their results on the Ubuntu corpus are reported, otherwise we implemented the models by tunning their parameters on the validation sets. All models were implemented using the Theano framework  \cite{2016arXiv160502688short}. Word embeddings in neural networks were  initialized by the results of word2vec \cite{mikolov2013distributed} \footnote{\url{https://code.google.com/archive/p/word2vec/}} pre-trained on the training data. We did not use Glove \cite{pennington2014glove} because the Ubuntu corpus contains many technical words that are not covered by Twitter or Wikipedia. The word embedding size was chosen as $200$. The maximum utterance length was set as $50$.
The maximum context length (i.e., number of utterances per context) was varied from $1$ to $20$ and set as $10$ at last. We padded zeros if the  number of utterances in a context is less than $10$, otherwise we kept the last $10$ utterances. We will discuss how performance of models changes in terms of different maximum context length later.

For SCN, the window size of convolution and pooling was tuned in $\{(2,2),(3,3)(4,4)\}$ and was set as $(3,3)$ finally. The number of feature maps is $8$. The size of the hidden states in the construction of $\mathbf{M}_2$ is the same with the word embedding size, and the size of the output vector $v$ was set as $50$. Furthermore, the size of the hidden states in the matching accumulation module is also $50$.	In SAN, the size of the hidden states in the segment level representation is $200$, and the size of the hidden states in Equation (\ref{attention}) was set as $400$. 
	
	All tuning was done according to $R_{2}@1$ on the validation data.

	\subsection{Evaluation Results}
	\begin{table*}[h]
		
		\centering
		\caption{Evaluation results on the Ubuntu corpus. Subscripts including ``last'', ``static'', and ``dynamic'' indicate three approaches to predicting a matching score as described in Section \ref{sec:matching}. Numbers in bold mean that the improvement from the models is statistically significant over the best baseline method. 
		}		\label{exp:response} 	
		
		\begin{tabular}{l|c|c|c|c}
			\thickhline
			&  $R_2@1$      &  $R_{10}@1$ &  $R_{10}@2$ &  $R_{10}@5$ \\ \hline
			TF-IDF  & 0.659 & 0.410 & 0.545 & 0.708 \\ 
			RNN  & 0.768 & 0.403 & 0.547 & 0.819\\ 
			CNN & 0.848 & 0.549 & 0.684 & 0.896\\ 
			LSTM & 0.901 & 0.638 & 0.784 & 0.949\\ 
			BiLSTM & 0.895 & 0.630 & 0.780 & 0.944 \\ \hline
			Multi-View  & 0.908 & 0.662 & 0.801 & 0.951\\ 
			DL2R  & 0.899& 0.626 & 0.783 & 0.944\\ \hline
			MV-LSTM & 0.906& 0.653 & 0.804 & 0.946 \\ 
			Match-LSTM & 0.904& 0.653 & 0.799 & 0.944 \\ 
			Attentive-LSTM & 0.903& 0.633 & 0.789 & 0.943 \\ 
			SCN$_{single}$ & 0.904& 0.656 & 0.809 & 0.942 \\  
			SAN$_{single}$& 0.906& 0.662 & 0.810 & 0.945\\ \hline
			SCN$_{last}$ & \textbf{0.923} & \textbf{0.723} & \textbf{0.842} & \textbf{0.956} \\
			SCN$_{static}$ & \textbf{0.927} & \textbf{0.725} & \textbf{0.838} & \textbf{0.962} \\
			SCN$_{dynamic}$ & \textbf{0.926} & \textbf{0.726} & \textbf{0.847} & \textbf{0.961} \\ \hline
			
			SAN$_{last}$ & \textbf{0.930} & \textbf{0.733} & \textbf{0.850} & \textbf{0.961} \\
			SAN$_{static}$ & \textbf{0.932} & \textbf{0.734} & \textbf{0.852} & \textbf{0.962} \\
			SAN$_{dynamic}$ & \textbf{0.932} & \textbf{0.733} & \textbf{0.851} & \textbf{0.961} \\
			\thickhline
		\end{tabular}	
	\end{table*}
	
	\begin{table*}[h]
		
		\centering
		\caption{Evaluation results on the Douban corpus. Notations have the same meaning with those in Table \ref{exp:response}. On $R_{10}@5$, only SAN significantly outperforms baseline methods.
		}		\label{exp:chinese_response} 	
		
		\begin{tabular}{l|c|c|c|c|c|c}
			\thickhline
			 & MAP & MRR & P@1 &  $R_{10}@1$ &  $R_{10}@2$ &  $R_{10}@5$\\ \hline
			TF-IDF & 0.331 &0.359 &0.180 & 0.096&0.172& 0.405\\ 
			RNN  & 0.390 &0.422 &0.208&0.118&0.223&0.589 \\ 
			CNN &  0.417 &0.440 &0.226&0.121&0.252&0.647\\ 
			LSTM &  0.485 & 0.527 &0.320&0.187&0.343&0.720\\ 
			BiLSTM &0.479&0.514&0.313&0.184&0.330&0.716\\ \hline
			Multi-View  &0.505&0.543&0.342&0.202&0.350&0.729\\ 
			DL2R  &0.488&0.527&0.330&0.193&0.342&0.705\\ \hline
			MV-LSTM & 0.498 & 0.538 & 0.348 &0.202&0.351&0.710 \\ 
			Match-LSTM &0.500& 0.537 & 0.345&0.202&0.348&0.720 \\ 
			Attentive-LSTM & 0.495& 0.523 & 0.331&0.192&0.328&0.718 \\ 
			SCN$_{single}$ & 0.506 & 0.543 & 0.349 &0.203&0.351&0.709 \\  
			SAN$_{single}$& 0.508 & 0.547 & 0.352 &0.206&0.353&0.720\\ \hline
			SCN$_{last}$ &\textbf{0.526}&\textbf{0.571}&\textbf{0.393}&\textbf{0.236}&\textbf{0.387}&0.729 \\
			SCN$_{static}$&\textbf{0.523}&\textbf{0.572}& \textbf{0.387}&\textbf{0.228}&\textbf{0.387}&0.734 \\
			SCN$_{dynamic}$ &\textbf{0.529}&\textbf{0.569}&\textbf{0.397}&\textbf{0.233}&\textbf{0.396}&0.724\\ \hline
			
			SAN$_{last}$	&\textbf{0.536}&\textbf{0.581}&\textbf{0.393}&\textbf{0.236}&\textbf{0.404}& \textbf{0.761} \\
			SAN$_{static}$ &\textbf{0.532}&\textbf{0.575}&\textbf{0.387}&\textbf{0.228}&\textbf{0.393}& \textbf{0.736}  \\
			SAN$_{dynamic}$ &\textbf{0.534}&\textbf{0.577}&\textbf{0.391}&\textbf{0.230}&\textbf{0.393}& \textbf{0.742}   \\
			\thickhline
		\end{tabular}	
	\end{table*}
	Table \ref{exp:response} and Table \ref{exp:chinese_response} show the evaluation results on the Ubuntu Corpus and the Douban Corpus respectively. SAN and SCN outperform baselines over all metrics on both data sets with large margins, and except $R_{10}@5$ of SCN on the Douban corpus, the improvements are statistically significant (t-test with $p$-value $\leq 0.01$). Our models are better than state-of-the-art single turn matching models such as MV-LSTM, Match-LSTM, SCN$_{single}$, and SAN$_{single}$. The results demonstrate that one cannot neglect utterance relationships and simply perform multi-turn response selection by concatenating utterances together. 
	
    TF-IDF shows the worst performance, indicating that the multi-turn response selection problem cannot be addressed with shallow features. LSTM is the best model among the basic models. The reason might be that it models relationships among words. Multi-View is better than LSTM, demonstrating the effectiveness of the utterance-view in context modeling. Advanced models have better performance, because they are capable of capturing more complicated structures in contexts.

	SAN is better than SCN on both data sets, which might be attributed to three reasons. The first reason is that SAN uses vectors instead of scalars to represent interactions between words or text segments. Therefore,  the matching vectors in SAN can encode more information from the pairs than those in SCN. The second reason is that SAN uses a soft attention mechanism to emphasize important words or segments in utterances, while SCN employs a max pooling operation to select important information from similarity matrices. When multiple words or segments are important in an utterance-response pair, a max pooling operation just selects the top one but the attention mechanism can leverage all of them. The last reason is that SAN models the sequential relationship and dependency among words or segments in the interaction aggregation module, while SCN only considers n-grams.
	
	The three approaches to matching prediction do not show much difference in both SCN and SAN, but dynamic average and static average are better than last state on the Ubuntu corpus and worse than it on the Douban corpus. This is because contexts in the Ubuntu corpus are longer than those in the Douban corpus (average context length $10.1$ v.s. $6.7$), and thus the last hidden state may lose information in history on the Ubuntu data. In contrast, the Douban corpus has shorter contexts but longer utterances (average utterance length $18.5$ vs $12.4$), and thus noise may be involved in response selection if more hidden states are taken into consideration. 
	
	There are two reasons that $R_n@k$s on the Douban corpus are much smaller than those on the Ubuntu corpus. One is that response candidates in the Douban corpus are returned by a search engine instead of negative sampling, which makes the problem harder. The other is that there are multiple correct candidates for a context, so the maximum  $R_{10}@1$ for some contexts are not $1$. For example, if there are $3$ correct responses, then the maximum $R_{10}@1$ is $0.33$.  P$@1$ is about 40$\%$ on the Douban corpus, indicating the difficulty of the task in a real chatbot.

	\subsection{Further Analysis}\label{analysis}
		\begin{table*}[h]  \small
		
		\centering
		\caption{Evaluation results of model ablation. \label{exp:discuss}}	
		\scalebox{0.85}{
			\begin{tabular}{l|c|c|c|c|c|c|c|c|c|c}
				\thickhline &   \multicolumn{4}{c|}{\textbf{Ubuntu Corpus}}    &        \multicolumn{6}{c}{\textbf{Douban Corpus}}        \\ \hline
				&  $R_2@1$      &  $R_{10}@1$ &  $R_{10}@2$ &  $R_{10}@5$ &MAP&MRR&P@1 &  $R_{10}@1$ &  $R_{10}@2$ &  $R_{10}@5$  \\ \hline
				Replace$_M$ & 0.905 & 0.661 & 0.799 & 0.950 & 0.503 &0.541 &0.343&0.201&0.364&0.729\\  \hline
			
				SCN with words & 0.919 & 0.704 & 0.832 & 0.955 & 0.518 &0.562 &0.370&0.228&0.371&0.737\\ 
				SCN with segments& 0.921 & 0.715 & 0.836 & 0.956 & 0.521 & 0.565 &0.382&0.232&0.380&0.734\\ 
				SCN	Replace$_A$ & 0.918 & 0.716 & 0.832 & 0.954 & 0.522&0.565 &0.376&0.220&0.385&0.727\\ 
				SCN$_{last}$ & 0.923 & 0.723 & 0.842 & 0.956 &0.526&0.571&0.393&0.236&0.387&0.729\\ \hline			
				SAN with words& 0.922 & 0.713 & 0.842 & 0.957 & 0.523 &0.565 &0.372&0.232&0.381&0.747\\ 
				SAN  with segments& 0.928 & 0.729 & 0.846 & 0.959 & 0.532 & 0.575 &0.385&0.234&0.393&0.754\\
				SAN	Replace$_A$ & 0.927 & 0.728 & 0.842 & 0.959 & 0.532&0.561 &0.386&0.225&0.395&0.757\\
				SAN$_{last}$ & 0.930 & 0.733 & 0.850 & 0.961 &0.536&0.581&0.393&0.236&0.404&0.761\\ \hline
			\end{tabular}
		}
	\end{table*}
	\subsubsection{Model ablation}	
	 We first investigated how different parts of SCN and SAN affect their performance by ablating SCN$_{last}$ and SAN$_{last}$.  Table \ref{exp:discuss} reports the results of ablation on the test data. First, we replaced the utterance-response matching module in SCN and SAN with a neural tensor \cite{socher2013reasoning} (denoted as Replace$_M$) which matches an utterance and a response by feeding their representations to a neural tensor network (NTN). The result is that the performance of the two models dropped dramatically. This is because in NTN, there is no interaction between the utterance and the response before their matching; and it is doubtful if NTN can recognize important parts in the pair and encode the information into matching. As a result, the model loses important information in the pair. Therefore, we can conclude that a good utterance-response matching mechanism is crucial to the success of SMF. At least, one has to let an utterance and a response interact with each other and explicitly highlight important parts in their matching vector. Second, we replaced the GRU in the matching accumulation modules of SCN and SAN with a multi-layer perceptron (MLP) (denoted as SCN Replace$_A$ and SAN Replace$_A$ respectively). The change led to a slight performance drop. This indicates that utterance relationships are useful in context-response matching. Finally, we only left one level of granularity, either word level or segment level, in SCN and SAN, and denoted the models as SCN with words, SCN with segments, SAN with words, and SAN with segments respectively. The results indicate that segment level matching on utterance-response pairs contributes more to the final context-response matching, and both segments and words are useful in response selection.

	\subsubsection{Comparison with respect to context length} 
\begin{figure}[t] \centering 
	\includegraphics[width=14cm,height=8cm]{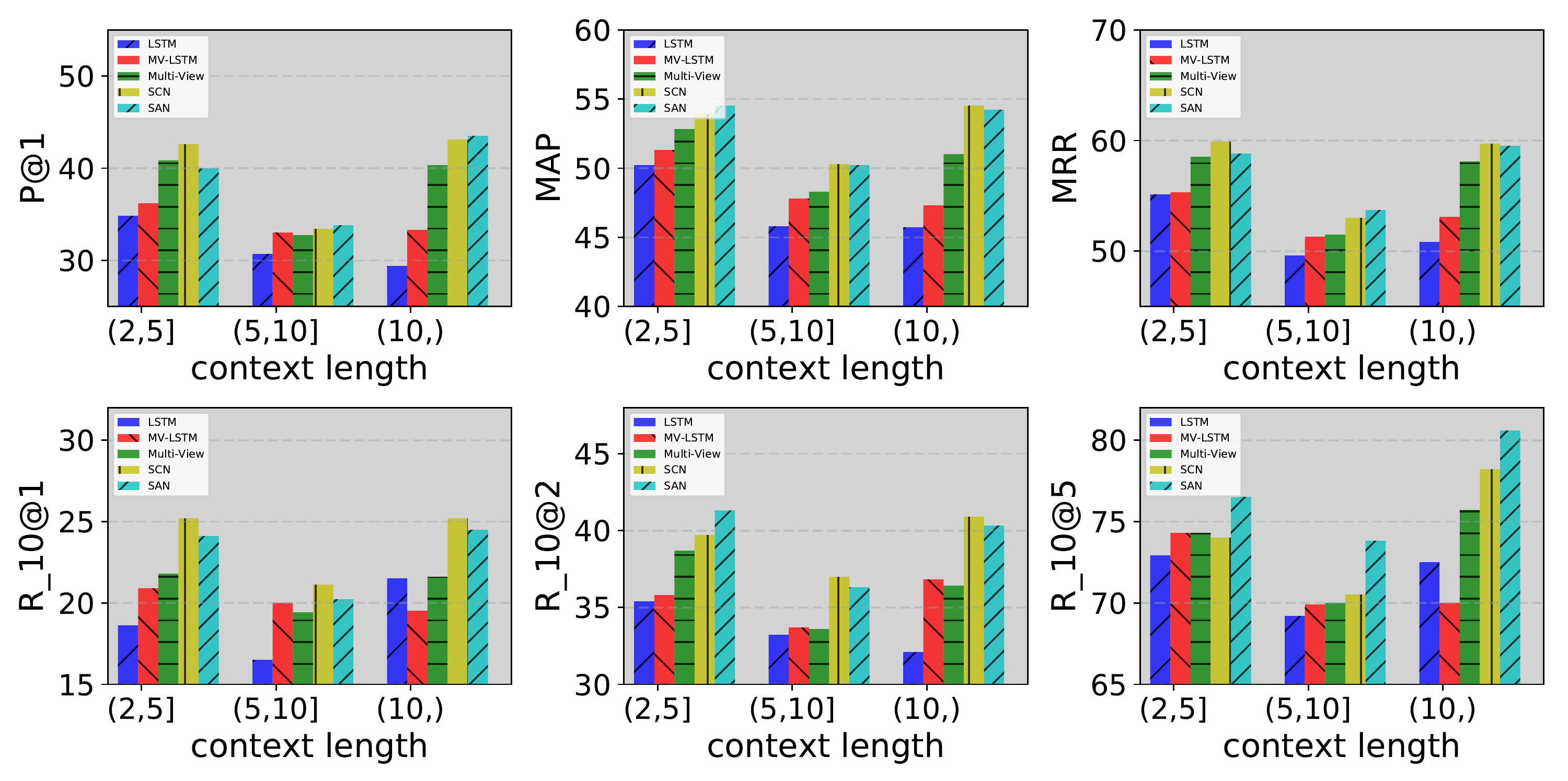}			\caption{Model performance across context length. We compared SAN and SCN with LSTM, MV-LSTM and Multi-View on the Douban corpus.}	\label{length}	
\end{figure}

We then studied how the performance of SCN$_{last}$ and SAN$_{last}$ changes across  contexts with different lengths. Context-response pairs were bucketed into $3$ bins according to the length of the contexts (i.e., the number of utterances in the contexts), and comparison was made in different bins on different metrics. Figure \ref{length} gives the results. Note that we did the analysis only on the Douban corpus because on the Ubuntu corpus many results were copied from the existing literatures and the bin-level results are not available.  SAN and SCN consistently perform better than the baselines over bins, and a general trend is that when contexts become longer, gaps become larger. For example, in $(2,5]$, SAN is $3$ points higher than LSTM on $R_{10}@5$, but the gap becomes $6$ points in $(10,)$. The results demonstrate that our models can well capture dependencies, especially long-distance dependencies, among utterances in contexts. SAN and SCN have similar trends because both of them use a GRU in the second layer to model dependencies among utterances.  

	\subsubsection{Sensitivity to hyper-parameters} We checked how sensitive SCN and SAN are regarding to the size of word embedding and the maximum context length.  Table \ref{wordembedding} reporsts evaluation results of SCN$_{last}$ and SAN$_{last}$ with embedding sizes varying in $\{50,100,200\}$. We can see that SAN is more sensitive to the word embedding size than SCN. SCN gets stable after the embedding size exceeds $100$, while SAN keeps getting improved with the increase of the embedding size.  Our explanation to the phenomenon is that SCN transforms word vectors and hidden vectors of GRU to scalars in the similarity matrices by dot products, and thus information in extra dimensions (e.g., entries with indices larger than $100$) might be lost; on the other hand, SAN leverages the whole $d$-dimensional vectors in matching, so the information in the embedding can be exploited more sufficiently.
	
\begin{figure*}[h]
	\centering
	\subfigure[Performance of SCN across different context length]{
		\includegraphics[width=13cm,height=6cm]{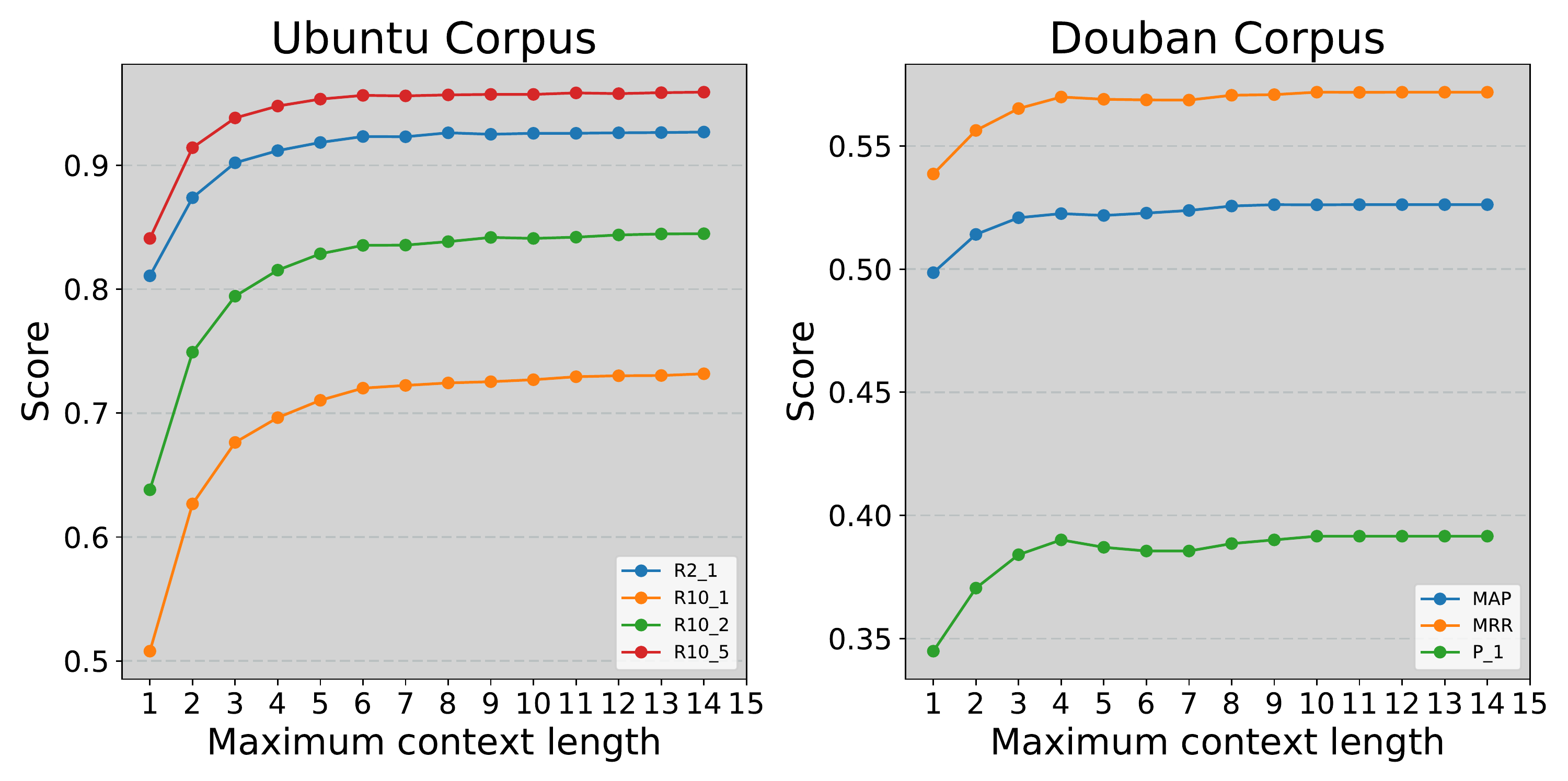}
	}
	\subfigure[Performance of SAN across different context length]{
		\includegraphics[width=13cm,height=6cm]{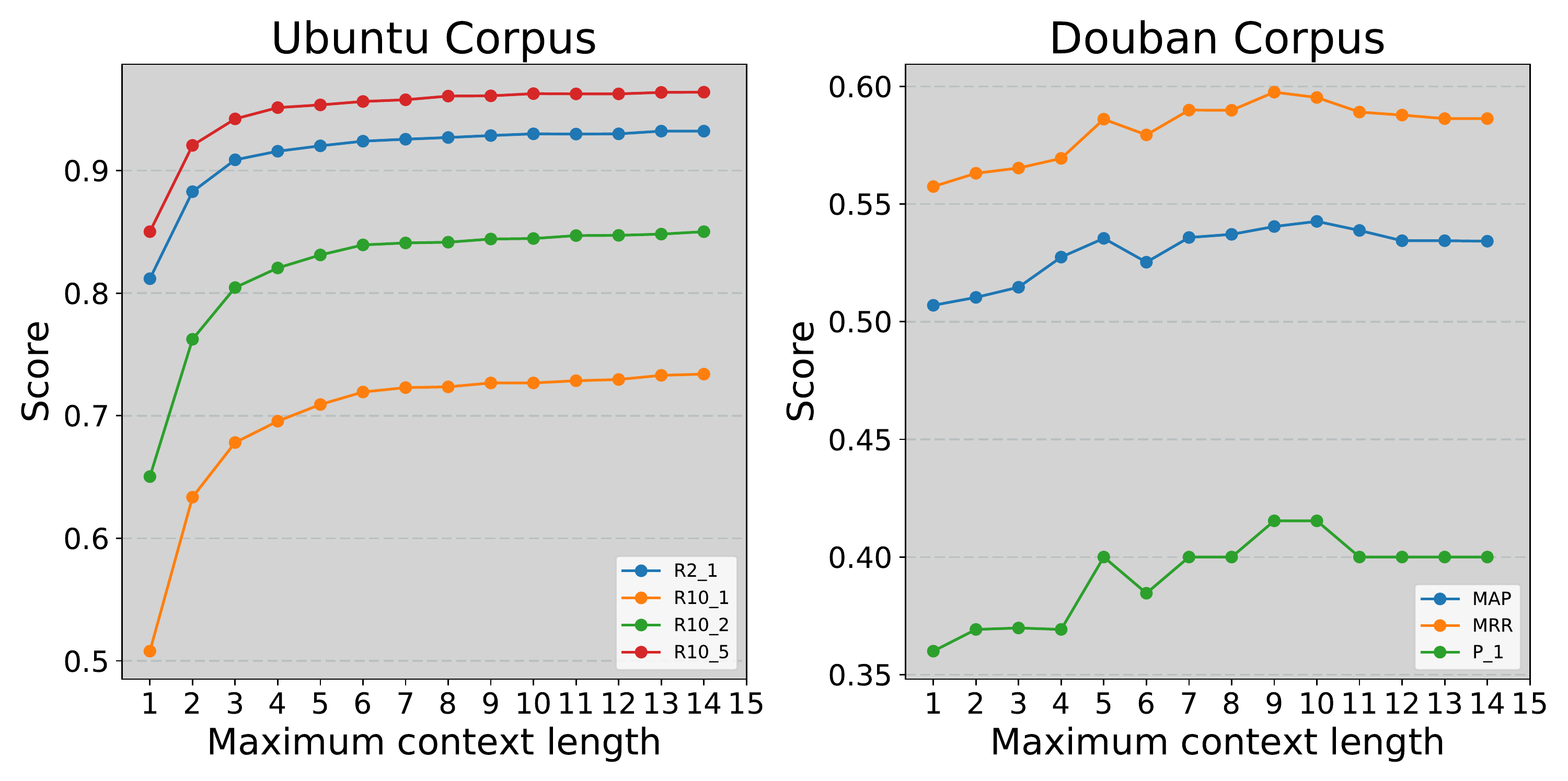}
	}
	\caption{Performance with respect to different maximum context length} \label{fig:max_smn_length}
\end{figure*}

	Figure \ref{fig:max_smn_length} gives the performance of SCN and SAN with respect to the maximum context length. We find that both models significantly become better with the increase of maximum context length when it is lower than $5$, and become stable after the maximum context length reaches $10$. The results indicate that utterances from early history can provide useful information to response selection.  Moreover, model performance  is more sensitive to the maximum context length on the Ubuntu corpus than it is on the Douban corpus. This is because utterances in the Douban corpus are longer than those in the Ubuntu corpus (average length $18.5$ v.s. $12.4$), which means single utterances in the Douban corpus could contain more information than those in the Ubuntu corpus. In practice,  we set the maximum context length as $10$ to balance effectiveness and efficiency.

	\begin{table*}[t]  \small
		
		\centering
		\caption{Evaluation results in terms of different word embedding sizes. \label{wordembedding} }	
		\scalebox{0.9}{
		\begin{tabular}{l|c|c|c|c|c|c|c|c|c|c}
			\thickhline &   \multicolumn{4}{c|}{\textbf{Ubuntu Corpus}}    &        \multicolumn{6}{c}{\textbf{Douban Corpus}}        \\ \hline
			&  R$_2$@1      &  R$_{10}$@1 &  R$_{10}$@2&  R$_{10}$@5 &MAP&MRR&P@1 &  R$_{10}$@1 &  R$_{10}$@2&  R$_{10}$@5  \\ \hline
			SCN$_{50d}$ & 0.920 & 0.715 & 0.834 & 0.952 & 0.503 &0.541 &0.343&0.201&0.364&0.729\\ 
			SCN$_{100d}$  & 0.921 & 0.718 & 0.838 & 0.954 & 0.524&0.569 &0.391&0.234&0.387&0.727\\ 
			SCN$_{200d}$ & 0.923 & 0.723 & 0.842 & 0.956 & 0.526 &0.571 &0.393&0.236&0.387&0.729\\ \hline
			SAN$_{50d}$ & 0.914 & 0.698 & 0.828 & 0.950 & 0.503 &0.541 &0.343&0.201&0.364&0.729\\ 
			SAN$_{100d}$  & 0.921 & 0.711 & 0.840 & 0.953 & 0.525&0.565 &0.375&0.220&0.388&0.746\\ 
			SAN$_{200d}$ & 0.930 & 0.733 & 0.850 & 0.961 &0.536&0.581&0.393&0.236&0.404&0.761\\ \hline
			
		\end{tabular}
	}
	\end{table*}

   \subsubsection{Model efficiency} 
\begin{figure}[h] \centering 
		\includegraphics[width=12cm,height=5cm]{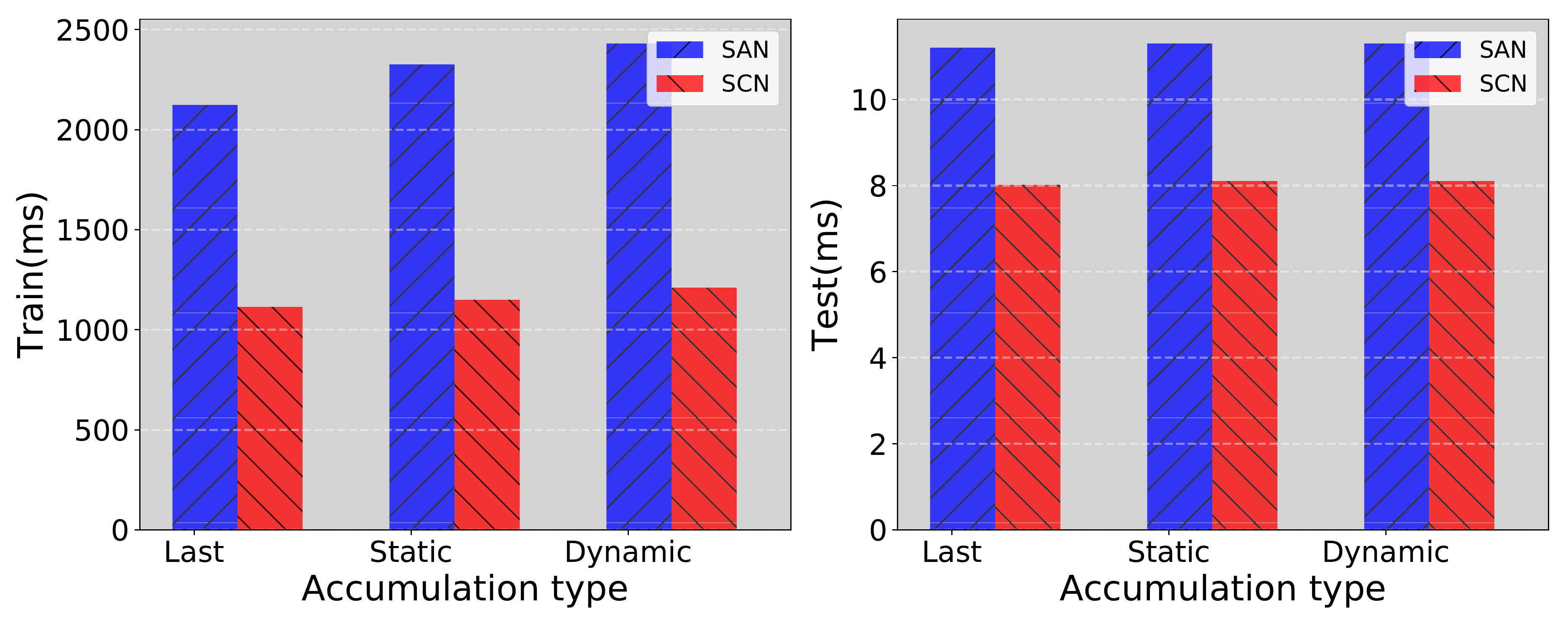}			\caption{Efficiency of SCN and SAN. The left figure shows the training time per batch with $200$ dimensional word embeddings, and the right one shows the inference time per batch. One batch contains $200$ instances.  } \label{efficiency}		
	\end{figure}

In Section \ref{section:vs}, we theoretically analyzed the efficiency of SCN and SAN. To verify the theoretical results, we further empirically compared their efficiency using the training data and the test data of the two data sets. The experiments were conducted using Theano on a Tesla K80 GPU with a Windows Server 2012 operation system. The parameters of the two models are described in Section \ref{parameter}. Figure \ref{efficiency} gives the training time and the test time of SAN and SCN. We can see that SCN is twice as fast as SAN in the training process, and saves $3$ millisecond per batch in the test process. Moreover, different matching functions do not influence the running time so much, as the bottleneck is the utterance representation learning.
	
	The empirical results are consistent with our theoreical results: SCN is faster than SAN. The results indicate that SCN is suitable for systems which care more about efficiency, whereas SAN can reach a higher accuracy with a little sacrifice of efficiency. 
	
    \subsubsection{Visualization}
	We finaly explained how SAN and SCN understand semantics of conversation contexts by visualizing the similarity matrices of SCN, the attnetion weights of SAN, and the update gate and the reset gate of the accumulation GRU of the two models using an example from the Ubuntu corpus. Table \ref{ubuntu_example} shows the example which is selected from the test set of the Ubuntu corpus and ranked at the top position by both SAN and SCN.
	
	\begin{table}[h]
		\caption{An example for visualization from the Ubuntu corpus \label{ubuntu_example}}	
		\centering
		\begin{tabular}{l}
			\hline
			\textbf{Context} \\ \hline
			\emph{u$_1$}: how can unzip many rar files at once? \\ \hline
			\emph{u$_2$}: sure you can do that in bash\\ \hline
			\emph{u$_3$}: okay how? \\ \hline
			\emph{u$_4$}: are the files all in the same directory?  \\ \hline
			
			\emph{u$_5$}: yes they all are;   \\ \hline
			\textbf{Response } \\ \hline
			\emph{Response}: then the command glebihan should extract them all from/to that directory \\ \hline
			
		\end{tabular}		
	\end{table}
	
\begin{figure*}[h] 
		\centering
		
		\subfigure[Visualization of M$_1$ in SCN. Darker squares refer to higher values.]{
				\includegraphics[width=12cm,height=6cm]  {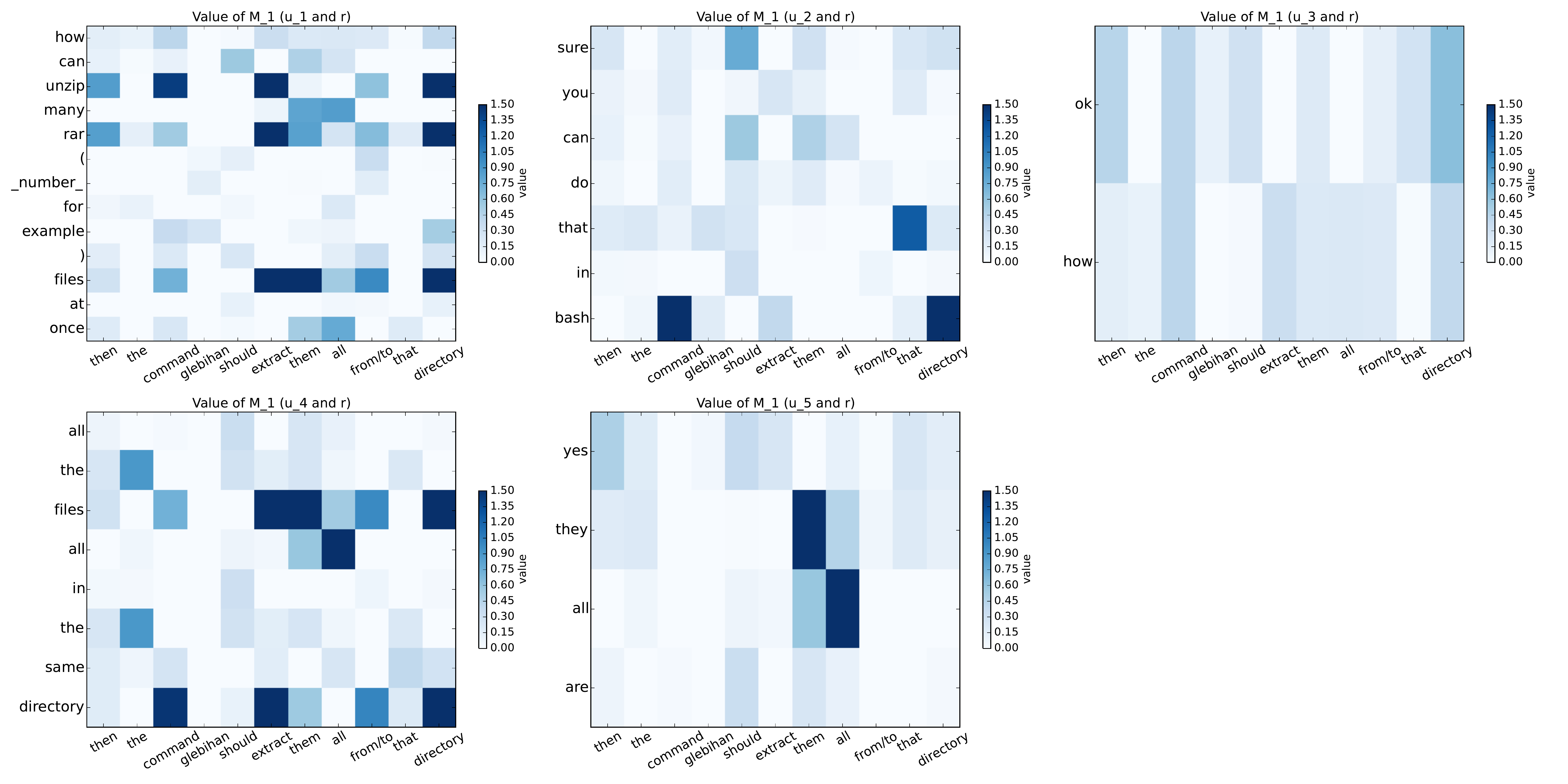}\label{scn_visual_m1}}		
		\subfigure[Visualization of M$_2$ in SCN. Darker sqaures refer to higher values.]{
			\includegraphics[width=12cm,height=6cm]{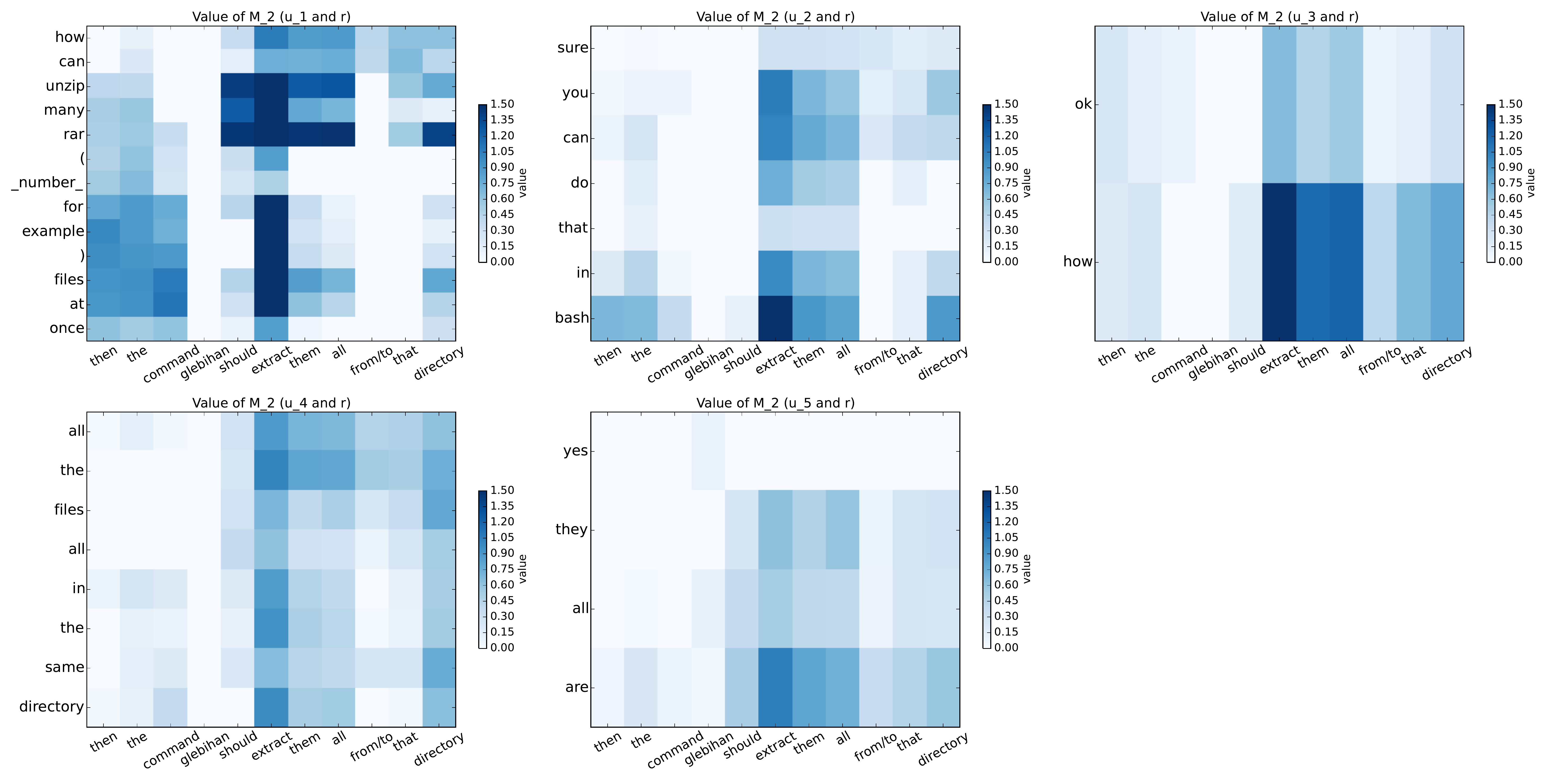}\label{scn_visual_m2}}	

		\subfigure[Visualization of gates. Darker squares refer to higher values.]{
				\includegraphics[width=12cm,height=4cm]{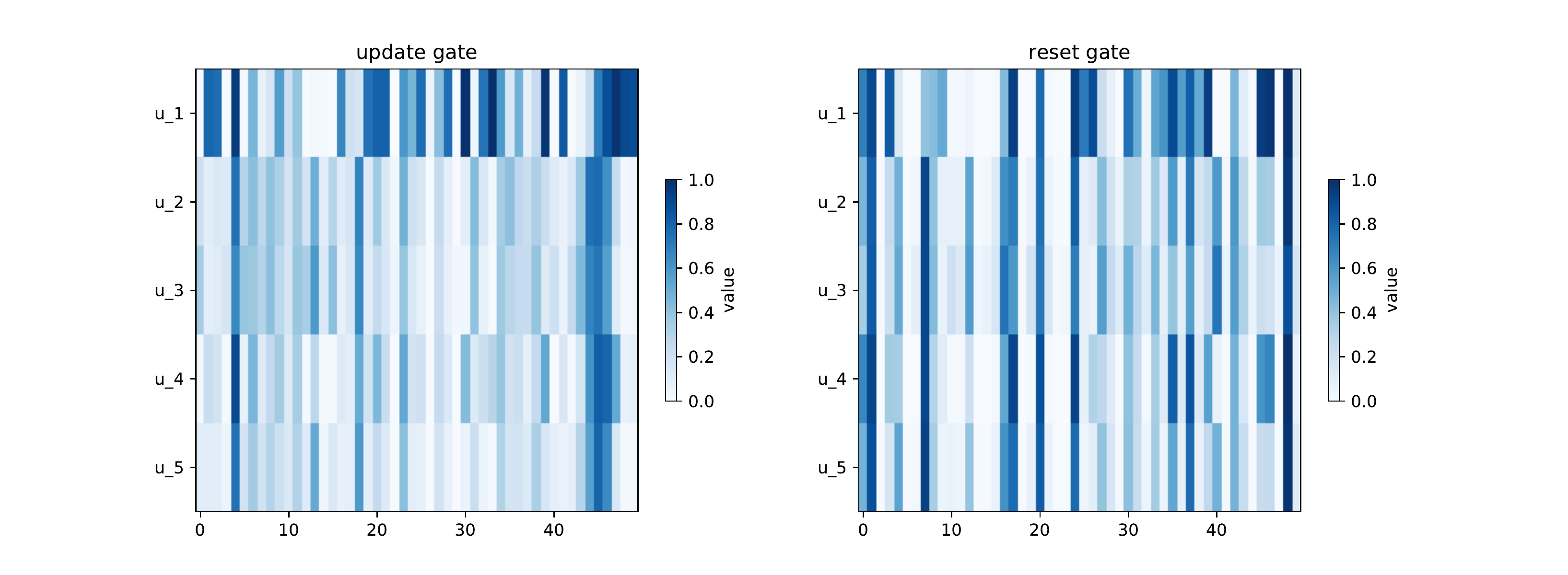} \label{scn_gates}}	
			\caption{Visualization of SCN}	\label{scn_visual}
	\end{figure*}

	
	Figure \ref{scn_visual_m1} illustrates word-word similarity matrices $\mathbf{M}_1$ in SCN. We can see that important words in $u_1$ such as ``unzip'', ``rar'', ``files'' are recognized and highlighted by words like ``command'', ``extract'', and ``directory'' in $r$. On the other hand, the similarity matrix of $r$ and $u_3$ is almost blank, as there is no important information conveyed by $u_3$. 
	Figure \ref{scn_visual_m2} shows the sequence-sequence similarity matrices $\mathbf{M}_2$ in SCN. We find that important segments like ``unzip many rar'' are highlighted, and the matrices also provide complementary matching information to  $\mathbf{M}_1$. Figure \ref{scn_gates} visualizes the reset gate and the update gate of the accumulation GRU respectively. Higher values in the update gate means that more information from the corresponding matching vector flows into matching accumulation. From Figure \ref{scn_gates}, we can see that $u_1$ is crucial to response selection and nearly all information from $u_1$ and $r$ flows to the hidden state of GRU, while other utterances are less informative and the corresponding gates are almost ``closed'' to keep the information from $u_1$ and $r$ until the final state.   
	
\begin{figure*}[h]
	\centering
	
	\subfigure[Visualization of A$_1$ in SAN. Darker squares refer to higher values.]{
		\includegraphics[width=12cm,height=6cm]{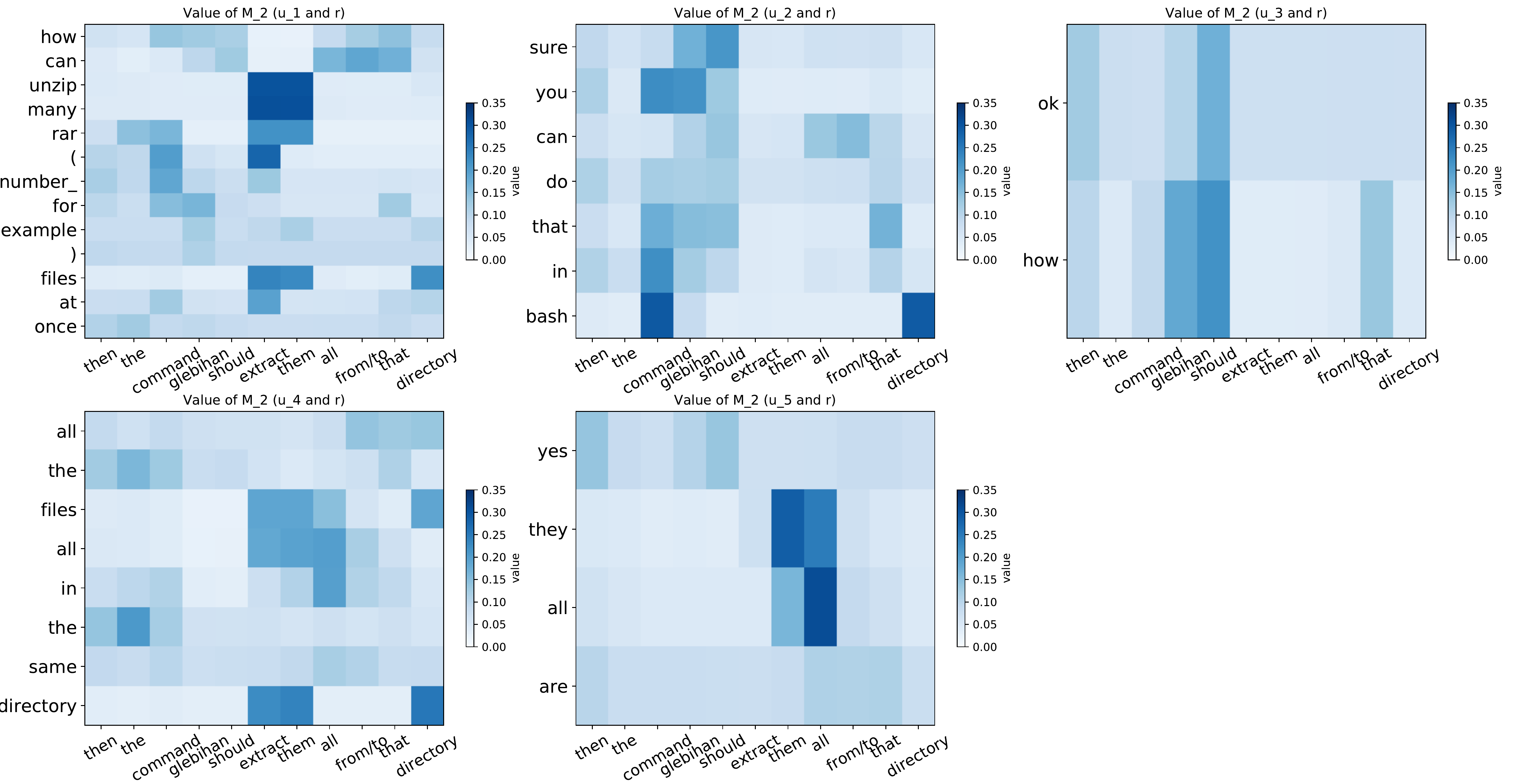}\label{san_visual_m1}}	
	\subfigure[Visualization of A$_2$ in SAN. Darker squares refer to higher values.]{
		\includegraphics[width=12cm,height=6cm]{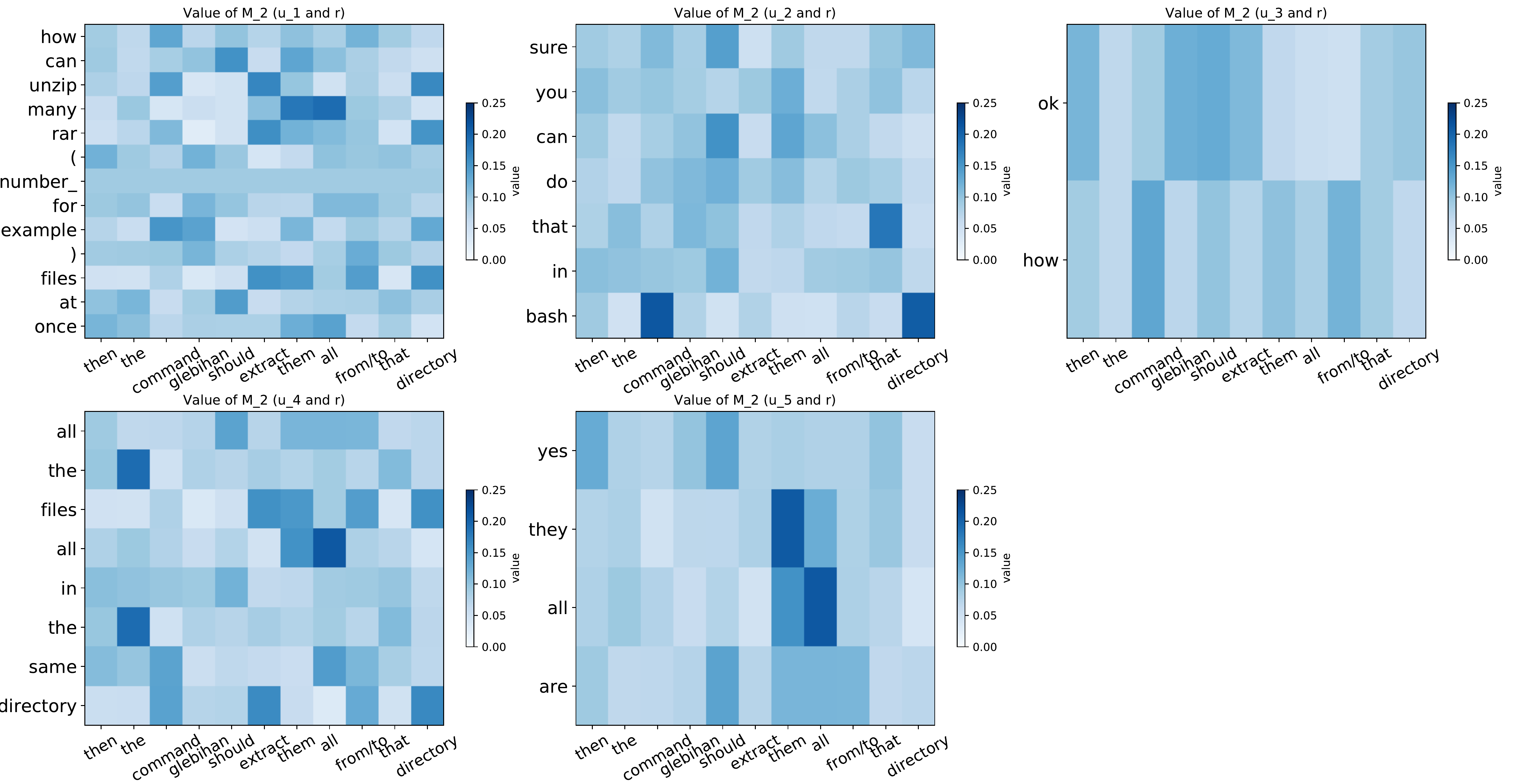}\label{san_visual_m2}}	
	
	\subfigure[Visualization of gates. Darker squares refer to higher values.]{
		\includegraphics[width=12cm,height=4.5cm]{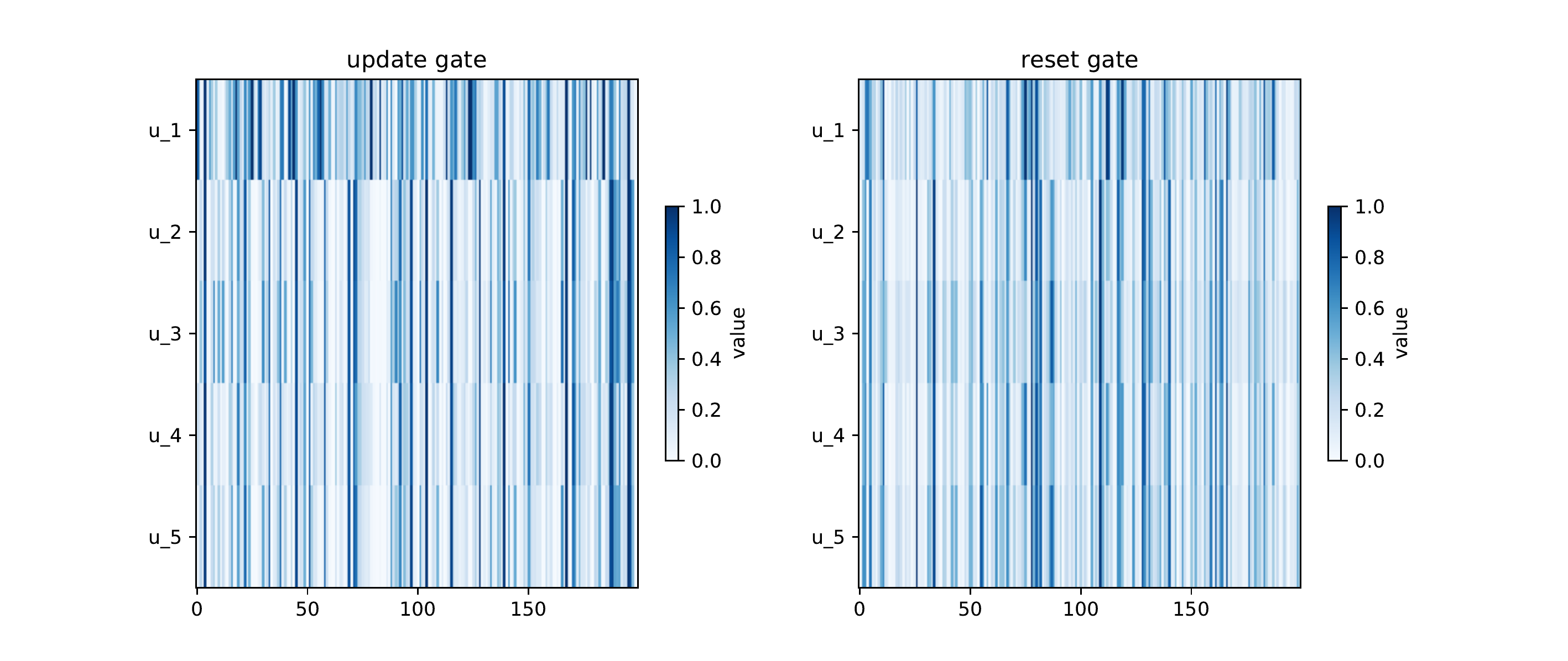}\label{san_visual_gates}}	
	\caption{Visualization of SAN}	\label{san_visual}
\end{figure*}

 	Regarding to SAN, Figure \ref{san_visual_m1} and Figure \ref{san_visual_m2} illustrate the word level attention weights $\mathbf{A_1}$ and segment level attention weights $\mathbf{A_2}$ respectively. Similar to SCN, important words such as ``zip" and ``file"  and important segments like ``unzip many rar'' get high weights, while function words like ``that" and ``for" are less attended. It should be noted that as the attention weights are normalized, the gaps between high and low values in  $\mathbf{A_1}$ and $\mathbf{A_2}$ are not so large as those in $\mathbf{M}_1$ and $\mathbf{M}_2$ of SCN. Figure \ref{san_visual_gates} visualizes the gates of the accumulation GRU, from which we observed similar distributions as those of SCN.

%
%

\subsection{Error analysis and future work}
	 Although models under SMF outperform baseline methods on the two data sets, there are still several problems that cannot be handled perfectly.
	
	(1) Logical consistency. SMF models the context and response on a semantic level, but pays little attention to logical consistency. This leads to several bad cases in the Douban corpus. We give a typical example in Table \ref{douban_example}. In the conversation history, one of the speakers says that he thinks the item on Taobao is fake , and the response is expected to say why he dislikes the fake shoes. However, both SCN and SAN rank the response `` It is not a fake. I just worry about the date of manufacture." at the top position. The response is inconsistent with the context on logic, as it claims that the jogging shoes are not fake which is contradictive to the context.
	
		\begin{table}[t]
		\caption{An example in the Douban corpus. The response is ranked at the top position among candidates, but it is inconsistent on logic to the current context. \label{douban_example}}	
		\centering
		\begin{tabular}{l}
			\hline
			\textbf{Context} \\ \hline
			\emph{u$_1$}: Does anyone know Newton jogging shoes?\\ \hline
			\emph{u$_2$}: 100 RMB on Taobao.\\ \hline
			\emph{u$_3$}: I know that. I do not want to buy it because that is a fake which is made in Qingdao , \\ \hline
			\emph{u$_4$}:Is it the only reason you do not want to buy it?  \\ \hline

			\textbf{Response } \\ \hline
			\emph{Response}: It is not a fake. I just worry about the date of manufacture. \\ \hline
			
		\end{tabular}		
	\end{table}
	The reason behind is that SMF only models semantics of context-response pairs. Logic, attitude and sentiment are not taken into account in response selection.
	
	  In the future, we shall explore the logic consistency problem in retrieval-based chatbots by leveraging more features.
	 
	(2) No valid candidates. Another serious issue is the  quality of candidates after retrieval. According to \cite{wu2016sequential}, the candidate retrieval method can be described as follows: given a message $u_n$ with $\{u_1,\ldots,u_{n-1}\}$ utterances in its previous turns, the top $5$ keywords are extracted from $\{u_1,\ldots,u_{n-1}\}$ based on their tf-idf scores\footnote{Tf is word frequency in the context, while idf is calculated using the entire index.}. $u_n$ is then expanded with the keywords, and the expanded message is sent to the index to retrieve response candidates using the inline retrieval algorithm of the index. The performance of the heuristic message expansion method is not good enough. In the experiment, only $667$ out of $1000$ contexts have correct candidates after response candidate retrieval. This indicates that there is still a big room to improve the retrieval component, and message expansion with several keywords from previous turns may not be enough for candidate retrieval. In the future, we will consider advanced methods for retrieving candidates.
	
	(3) Gap between training and test. Current method requires a huge amount of training data (i.e., context-response pairs) to learn a matching model. However, it is too expensive to obtain large scale (e.g., millions of) human labeled pairs in practice. Therefore, we regard conversations with human replies as positive instances and conversations with randomly sampled replies as negative instances in model training. This is a poor approximation to the real situation as we expect our models can distinguish positive responses from negative responses judged by humans. Because of the gap in training and test, our matching models, although perform much better than the baseline models, is still far from perfect on the Douban corpus (see the low P@1 in Table \ref{exp:chinese_response}). In the future, we may consider using small human labeled data but leveraging the large scale unlabeled data to learn matching models. 
		
		\newpage
	\section{Conclusion} \label{conclusion}
	In this paper, we study the problem of multi-turn response selection in which one has to model the relationships among utterances in a context and pay more attention to important parts of the context. We find that the existing models cannot address the two challenges at the same time when we summarize them into a general framework. Motivated by the analysis, we propose a sequential matching framework for context-response matching. The new framework is able to capture the important information in a context and model the utterance relationships simultaneously. Under the framework, we give two specific models based on a convolution-pooling technique and an attention mechanism. We test the two models on two public data sets. The results indicate that both models can significantly outperform the state-of-the-art models. To further understand the models, we conduct ablation analysis and visualize key compontents of the two models. We also compare the two models in terms of their efficacy, efficiency, and sensitivity to hyper-parameters.

	\starttwocolumn
	\bibliographystyle{compling}
	\bibliography{compling_style}
	
\end{document}